\documentclass[10pt,twocolumn,letterpaper]{article}

\usepackage[pagenumbers]{cvpr} 

\usepackage{graphicx}
\usepackage{balance}
\usepackage{amsmath}

\DeclareMathOperator*{\argmin}{arg\,min}

\usepackage{amssymb}
\usepackage{booktabs}
\usepackage{multirow, multicol}
\usepackage{makecell}
\usepackage[table,dvipsnames]{xcolor}

\usepackage[pagebackref,breaklinks=true,colorlinks,urlcolor=NavyBlue,bookmarks=false]{hyperref}
\usepackage[accsupp]{axessibility}  
\hypersetup{citecolor=[RGB]{119,185,0}}

\usepackage[capitalize]{cleveref}
\crefname{section}{Sec.}{Secs.}
\Crefname{section}{Section}{Sections}
\Crefname{table}{Table}{Tables}
\crefname{table}{Tab.}{Tabs.}

\def\cvprPaperID{8635} 
\def\confName{CVPR}
\def\confYear{2023}

\begin{document}

\title{DeepMapping2: Self-Supervised Large-Scale LiDAR Map Optimization}

\author{Chao Chen$^{1,}\thanks{Equal contribution}$,\, Xinhao Liu$^{1,}\footnotemark[1]$,\, Yiming Li$^{1}$,\, Li Ding$^{2}$,\, Chen Feng$^{1,}$\thanks{The corresponding author is Chen Feng {\tt\small cfeng@nyu.edu}}\\
$^{1}$New York University \quad $^{2}$ University of Rochester\\
{\tt\small \url{https://ai4ce.github.io/DeepMapping2/}}
}
\maketitle

\begin{abstract}
LiDAR mapping is important yet challenging in self-driving and mobile robotics.
To tackle such a global point cloud registration problem, DeepMapping~\cite{ding2019deepmapping} converts the complex map estimation into a self-supervised training of simple deep networks.
Despite its broad convergence range on small datasets, DeepMapping still cannot produce satisfactory results on large-scale datasets with thousands of frames. This is due to the lack of loop closures and exact cross-frame point correspondences, and the slow convergence of its global localization network. We propose DeepMapping2 by adding two novel techniques to address these issues: (1) organization of training batch based on map topology from loop closing, and (2) self-supervised local-to-global point consistency loss leveraging pairwise registration. Our experiments and ablation studies on public datasets such as KITTI, NCLT, and Nebula  demonstrate the effectiveness of our method.
\vspace{-8mm}
\end{abstract}
\providecommand{\titlevariable}{DeepMapping2}
\section{Introduction}
\label{sec:intro}


Mapping is a fundamental ability for autonomous mobile agents. It organizes an agent's local sensor observations into a map, i.e., a global spatial representation of the environment. A pre-built map is useful in robotics, self-driving, and augmented reality for agents to localize themselves~\cite{wolcott2014visual, lu2019l3, palieri2020locus, taguchi2013point, sarlin2022lamar}.
Various simultaneous localization and mapping (SLAM) methods can create maps of new environments from 2D and/or 3D sensors~\cite{davison2007monoslam,klein2007parallel, izadi2011kinectfusion,engel2014lsd,labbe2019rtab,campos2021orb,zhou2021lidar}.
In particular, LiDAR-based mapping is often adopted to build large-scale maps in self-driving and mobile robotics due to LiDAR's direct and accurate 3D point cloud sensing capability.


Similar to visual SLAM, LiDAR SLAM methods typically contain front-end and back-end modules~\cite{zhang2014loam,shan2018lego,shan2020lio,ebadi2020lamp}. The front-end module tracks sensor movements by LiDAR/inertial/wheel odometry and provides constraints between sequential frames of point clouds by either iterative closest point (ICP) or 3D feature detection and correspondence matching algorithms. The back-end uses those constraints in a pose/pose-landmark graph optimization~\cite{kaess2008isam,kummerle2011g} to minimize the odometry drift, similar to the bundle adjustment in visual SLAM and Structure-from-Motion (SfM).

However, without accurate GNSS/IMU as odometry, large-scale LiDAR mapping results could be unsatisfactory (see Fig.~\ref{fig:kitti_traj}), due to errors in LiDAR odometry and difficulties in correspondence matching and loop closing, especially outdoors. To tackle these issues, researchers start to explore deep learning methods. Some of them focus on replacing sub-tasks in LiDAR mapping with deep networks~\cite{yew20183dfeat,shi2021keypoint,li2019net,chen2021overlapnet}, following the common machine learning paradigm: \texttt{train-then-test}. Yet such methods could face generalization issues when the training dataset domain is different than the testing one.


Differently, DeepMapping~\cite{ding2019deepmapping} proposes a new paradigm: \texttt{training-as-optimization} for point cloud mapping. It encapsulates the global registration in a point-cloud-based PoseNet~\cite{kendall2015posenet} (L-Net), and evaluates the map quality using another binary occupancy network (M-Net) with a binary cross-entropy (BCE) loss. This converts the continuous map optimization into a self-supervised training of binary classifications. Since \textit{no testing is needed}, it does not face any generalization issues because mapping is done once training is finished.

However, despite its superior performance on small-scale datasets, we found DeepMapping often fails on large-scale datasets due to the following challenges:

(1) \textit{No-explicit-loop-closure}: DeepMapping gradually optimizes L-Net using frames in each mini-batch that are temporal neighbors, and only relies on M-Net to control the global map consistency. This is like incremental registration that is doomed to drift when the number of frames is large. SLAM solves this by loop closing, which is not yet clear how to be incorporated into DeepMapping.

(2) \textit{No-local-registration}: Although previous works have shown local registration to be locally accurate~\cite{yang2015go,singandhupe2021registration,wang2019deep,choy2020deep}, DeepMapping only uses it in the ICP-based pose initialization but not in the optimization. This is due to a common problem faced by all LiDAR registration methods, the lack of point correspondences in LiDAR point clouds: the same 3D point rarely appears again in another scan, because of the sparse sensor resolution and long-range sensing.

(3) \textit{Slow-convergence-in-global-registration}: L-Net regresses a single frame of point cloud into its global pose, which is supervised only by the M-Net and BCE loss. Unlike pairwise registration, this global registration lacks enough inference cues to output correct poses, thus leading to slow convergence when the dataset is large.

We propose \titlevariable~that is able to effectively optimize maps on large-scale LiDAR datasets. It extends DeepMapping with two novel techniques. The first one addresses challenge (1) by organizing data frames into training batches based on map topology from loop closing. This allows a frame with its topological/spatial neighbors to be grouped into the same batch. We find this to be the best way of adding loop closing into DeepMapping which uses free-space inconsistencies via M-Net and BCE loss to generate self-supervision, because such inconsistencies happen mostly between unregistered neighboring frames.

 The second technique is a novel self-supervised local-to-global point consistency loss that leverages precomputed pairwise registration. For each point in a frame, we can compute this new consistency as the L2 distance between different versions of its global coordinate calculated using a neighboring frame's global pose and the relative pose between the two frames from the pairwise registration. This allows us to address challenge (2) without relying on point correspondences between different frames: even if two neighboring frames do not have enough common points as correspondences for pairwise local registration, we can still incorporate the local registration's results during training. It also addresses the challenge (3) because now L-Net is supervised by stronger gradients from not only the BCE loss, but also the new consistency loss.
Our contributions are summarized as follows:
\begin{itemize}
    \vspace*{-1mm}
    \item Our \titlevariable~is the first self-supervised large-scale LiDAR map optimization method as far as we know, and this generic method achieves state-of-the-art mapping results on various indoor/outdoor public datasets, including KITTI\cite{geiger2013vision}, NCLT\cite{carlevaris2016university}, and the challenging underground dataset  Nebula\cite{agha2021nebula}. 
    \vspace*{-1mm}
    \item Our analysis reveals why DeepMapping fails to scale up and leads to the two novel techniques--batch organization and local-to-global point consistency loss--to incorporate loop closing and local registration in the DeepMapping framework. Their necessity and effectiveness are further validated in our ablation study.
    \vspace*{-1mm}
\end{itemize}
\section{Related Work}
\textbf{Pairwise registration.} Pairwise registration computes the transformation from a source point cloud to a target point cloud. A global pose of each point cloud can be obtained by incrementing the pairwise transformation sequentially. Iterative closest point (ICP)\cite{besl1992method} and its variants~\cite{yang2015go, maron2016point} are widely-used pairwise registration methods by iteratively estimating and minimizing closest point correspondence. More recently, learning-based algorithms use a pre-trained model to predict pairwise transformation~\cite{wang2019deep, choy2020deep}. However, all these methods suffer from aggregation errors when the registration is done incrementally to solve a mapping problem. Though \titlevariable~relies on pairwise registration for consistency loss, it only benefits from the relatively accurate pairwise transformation but does not suffer aggregation errors because registration is not done incrementally.


\textbf{Multiple registration.} In addition to pairwise registration, several methods for multiple point cloud registration have been proposed~\cite{theiler2015globally, evangelidis2014generative, izadi2011kinectfusion}. Global SfM~\cite{cui2015global, zhu2018very} divides one scene into smaller partitions and iteratively does motion averaging until convergence. \cite{choi2015robust} transforms the problem into pose-graph optimization and uses partitioning to reduce computation. \cite{gojcic2020learning} follows a similar practice but with a supervised end-to-end learnable algorithm. On the other hand, DeepMapping~\cite{ding2019deepmapping} does not require any scene partitioning or supervised training. Instead, it approaches the problem by transforming global map optimization into self-supervised training of deep networks. DeepMapping2 follows the same line of thought and inherits the general framework of DeepMapping. It improves the mapping result of DeepMapping, especially in large-scale scenes, and overcomes challenges faced by DeepMapping by the two proposed novel techniques.

\textbf{Loop closure.} Loop closure is a significant component in SLAM algorithms. It aims to decide whether a place has been visited before by the agent. Most classical loop closure methods~\cite{jegou2010aggregating} rely on hand-crafted features and descriptors to compare different frames. Deep learning approaches like PointNetVLAD~\cite{uy2018pointnetvlad} combine the feature extraction from PointNet~\cite{qi2017pointnet} and the supervised contrastive loss from NetVLAD~\cite{arandjelovic2016netvlad}. In contrast, TF-VPR~\cite{chen2022self} trains a similar deep-learning based network without supervision by mining the information between temporal and feature neighbors. It is a challenge, however, to incorporate loop closing in the training of DeepMapping because the detected loop closure cannot be described in a differentiable manner. DeepMapping2 solves this problem by organizing the training batches with spatially adjacent frames and incorporating loop closure information into the self-supervised training process.

\textbf{Simultaneous Localization and Mapping.} 
SLAM is the computational problem of building a map while keeping track of an agent's location. Both visual\cite{mur2015orb,mur2017orb,li2020deepslam} and LiDAR SLAM\cite{besl1992method,bosse2009keypoint, zlot2014efficient,shan2018lego,zhang2014loam} are well-studied in this domain. Particularly, LiDAR SLAM provides relatively accurate geometric information and can be classified into train-then-test and train-as-optimization. The majority of methods fall into the train-then-test category. Such LiDAR SLAM methods perform point-level matching\cite{besl1992method}, feature-level matching\cite{bosse2009keypoint, zlot2014efficient}, and point feature to edge/plane matching\cite{shan2018lego,zhang2014loam} to find correspondences between scans. However, existing LiDAR SLAM algorithms are prone to large errors, particularly in estimating the sensor rotation of each frame. DeepMapping\cite{ding2019deepmapping}, on the other hand, belongs to the train-as-optimization category, and the network trained does not generalize in other scenes because the self-supervised loss functions in \titlevariable~are designed for same scene optimization.


\providecommand{\mo}{\mathbf{o}}
\providecommand{\mq}{\mathbf{q}}
\providecommand{\lgloss}{consistency loss}
\newcommand{\lc}{\left ( }
\newcommand{\rc}{\right ) }

\section{Method}

\begin{figure*}[ht!]
    \centering
    \includegraphics[width=0.9\linewidth]{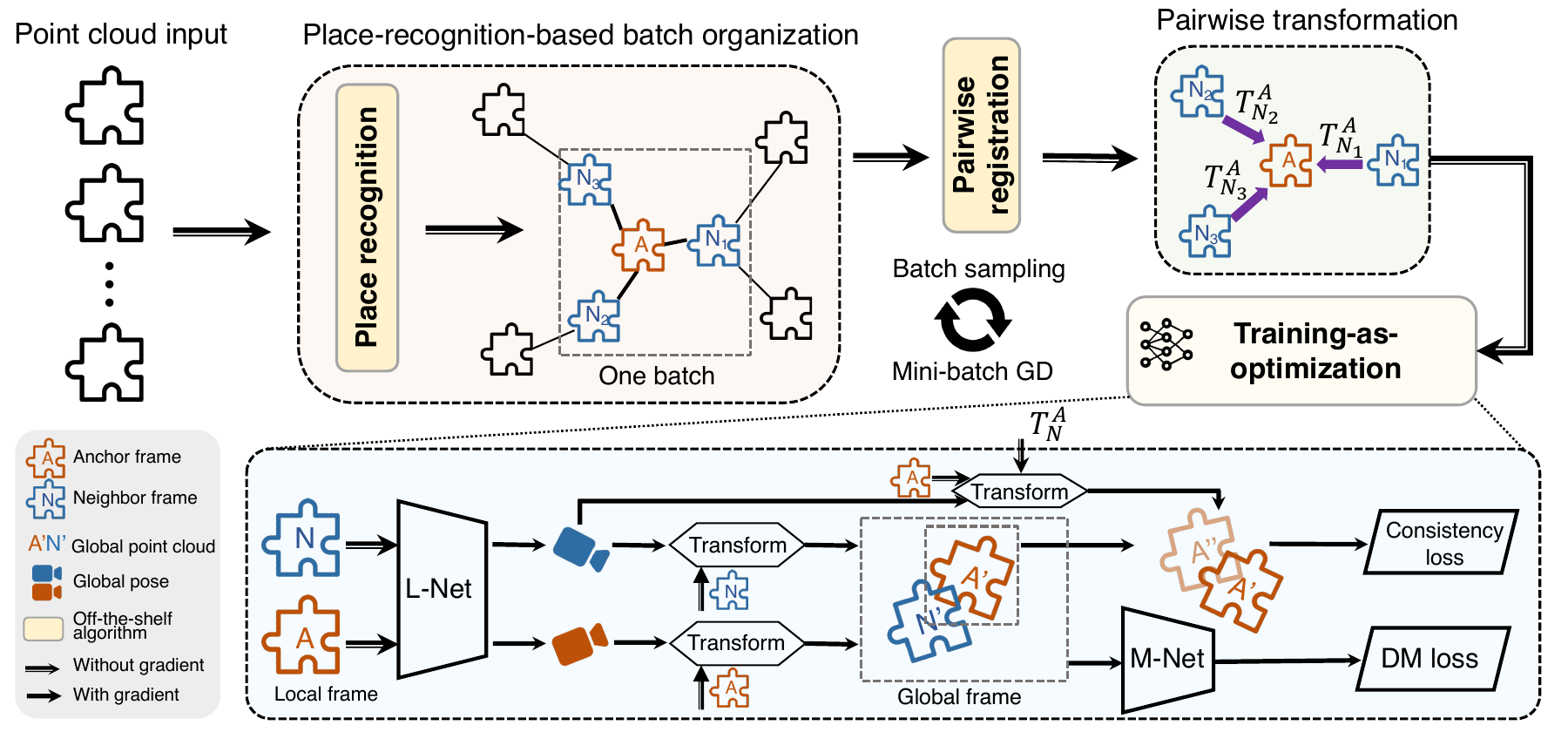}
    \caption{\textbf{Pipeline of \titlevariable.} The pipeline mainly consists of place-recognition-based \textit{batch organization} and learning-based \textit{optimization}. In batch organization, the input point clouds are organized into mini-batches by topological map attained from place recognition. Each batch contains an anchor frame $A$ and several spatially closed neighbor frames $N$. The transformation between the anchor frame and each neighbor frame $T^A_N$ is obtained by pairwise registration. In optimization, each batch is fed into L-Net to estimate the global pose. The transformed global anchor frame is then obtained in two ways: $A'$ directly from the global pose of the anchor frame ($T_A^G$) and $A''$ from the global pose of the neighbor frame ($T_N^G$) and the pairwise registration ($T^A_N$). The consistency loss penalizes the differences between $A'$ and $A''$. The global frames are also fed into M-Net for computing DeepMapping loss. Best viewed in color.}
    \label{fig:pipeline}
    \vspace{-4mm}
\end{figure*}

\subsection{Overview}\label{sec:overview}

\textbf{Problem setup.} 
We aim to solve the problem of registering multiple point clouds into a single global frame. Formally, the input point cloud is denoted as  $\mathcal{S}=\{S_i\}_{i=1}^K$, where $K$ is the total number of point clouds. Each point cloud $S_i$ is represented by a $N_i\times 3$ matrix where $N_i$ is the number of points in $S_i$, \ie$S_i\in\mathbb{R}^{N_i\times 3}$. The goal is to estimate the sensor pose  $\mathcal{T}=\{T_i\}_{i=1}^K$ for each $S_i$, where $T_i \in SE(3)$.


\textbf{Common ground with DeepMapping.}
Inspired by DeepMapping~\cite{ding2019deepmapping}, we use deep neural networks, represented as a function $f$ with learnable parameters, to estimate sensor poses $\mathcal{T}$. The registration problem is then transformed into finding the optimal network parameters that minimize the objective function
\vspace{-1mm}
\begin{equation}\label{eq:dm_loss}
    (\theta^*, \phi ^*) = \argmin_{\theta,\phi}\mathcal{L}_\phi(f_\theta(\mathcal{S}),\mathcal{S}) \textrm{,}
    \vspace{-1mm}
\end{equation}

where $f_\theta:S_i \mapsto T_i$ is a localization network (L-Net) that estimates the global pose for each point cloud, and $\phi$ is the parameter of a map network (M-Net) $m_\phi$ that maps from a global coordinate to its corresponding occupancy probability. 
$\mathcal{L}_\phi$ is the self-supervised binary cross entropy loss to measure the global registration quality:
\vspace{-1mm}
\begin{equation}\label{eq:MNet_loss}
\mathcal{L}_\phi=\frac{1}{K} \sum_{i=1}^{K} B\left[m_{\phi}\left(G_{i}\right), 1\right]+B\left[m_{\phi}\left(s\left(G_{i}\right)\right), 0\right]\textrm{,}
\end{equation}
\vspace{-1mm}
where the global point cloud $G_{i}$ is a function of L-Net parameters $\theta$, and $s(G_i)$ is a set of points sampled from the free space in $G_{i}$. $B[p, y]$ in \cref{eq:MNet_loss} is the binary cross entropy between the predicted occupancy probability $p$ and the self-supervised binary label $y$:

\vspace{-1mm}
\begin{equation}\label{eq:bce}
B[p, y]=-y \log (p)-(1-y) \log (1-p)\textrm{.}
\end{equation}

Moreover, Chamfer distance is another loss to help the network converge faster in DeepMapping~\cite{ding2019deepmapping}. It measures the distance of two global point clouds X and Y by:
\vspace{-1mm}
\begin{equation}\label{eq:chamfer}
\begin{aligned}
d(X, Y) &=\frac{1}{|X|} \sum_{\mathbf{x} \in X} \min _{\mathbf{y} \in Y}\|\mathbf{x}-\mathbf{y}\|_2 \\
&+\frac{1}{|Y|} \sum_{\mathbf{y} \in Y} \min _{\mathbf{x} \in X}\|\mathbf{x}-\mathbf{y}\|_2 \textrm{.}
\end{aligned}
\end{equation}

\textbf{Difference from DeepMapping.} DeepMapping introduces a "warm start" mechanism to transform each raw point cloud to a sub-optimal global pose via existing registration methods like ICP. This optional step will accelerate the convergence of DeepMapping in the small-scale dataset. However, a reasonable initialization is required in our method for large-scale datasets, because it would be difficult to converge if starting from scratch. 

\textbf{Limitation of DeepMapping.} Despite its success in small datasets, DeepMapping cannot scale up to large datasets because of the aforementioned challenges: (1) \textit{no-explicit-loop-closure}, (2) \textit{no-local-registration}, and (3) \textit{slow-convergence-in-global-registration}.

In \cref{sec:pipeline}, we will introduce the pipeline of our method. We will introduce the solution to challenge (1) in \cref{sec:batch}, and the solution to challenges (2) and (3) in \cref{sec:consistency}

\subsection{Pipeline}\label{sec:pipeline}
\textbf{Batch organization.} The pipeline of DeepMapping2 has two main stages as shown in yellow and blue blocks in \cref{fig:pipeline}. The first stage is the organization of training batches based on map topology from place recognition. The input is a sequence of point clouds that is initially registered to an intermediate state between scratch and the ground truth. Each anchor frame $A$ is organized with its neighbor frame $N$ into one batch using the map topology from off-the-shelf place recognition algorithms~\cite{uy2018pointnetvlad, chen2022self} ($A$ and $N$ are formally defined in \cref{sec:batch}).

\textbf{Pairwise registration.} Before the optimization in the second stage, pairwise registration is calculated between the anchor frame and the neighbor frames in each batch by finding the transformation from the neighbors to the anchor. This can also be achieved by any off-the-shelf pairwise registration algorithms. 

\textbf{Training as optimization.} The second stage of the pipeline is learning-based optimization. Besides those important components that have been introduced in \cref{sec:overview}, the \lgloss~is another key component of \titlevariable. The loss function is designed with the following idea: for each point in the anchor frame $A$, we can compute its consistency as the L2 difference between different versions ($A'$ and $A''$) of its global coordinates.  These versions are calculated from the global pose of each neighboring frame $N$ and the relative transformation between the anchor and the neighbor. A more detailed explanation of the loss function will be given in \cref{sec:consistency}.


\subsection{Batch organization}\label{sec:batch}

\begin{figure}
    \centering
    \begin{subfigure}{0.15\textwidth}
        \includegraphics[width=\textwidth]{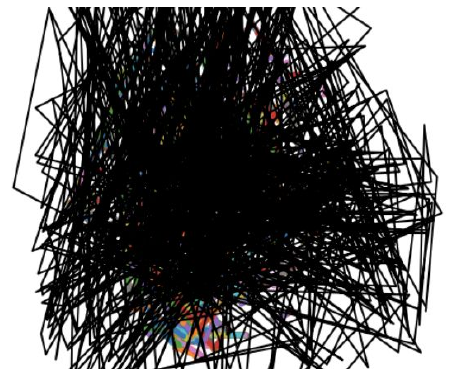}
        \caption{Random}
        \label{fig:batch1}
    \end{subfigure}
    \begin{subfigure}{0.15\textwidth}
        \includegraphics[width=\textwidth]{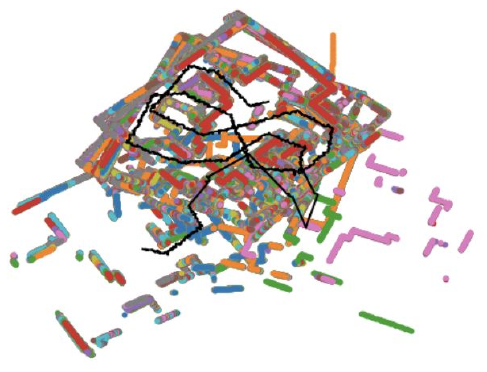}
        \caption{Temporal}
        \label{fig:batch2}
    \end{subfigure}
    \begin{subfigure}{0.15\textwidth}
        \includegraphics[width=\textwidth]{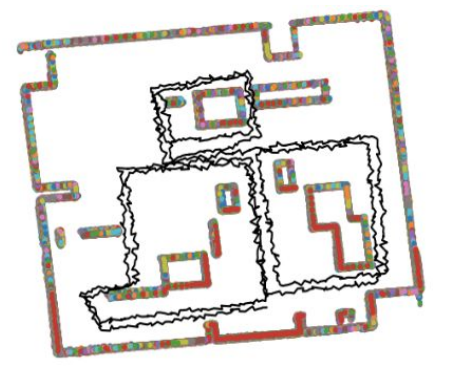}
        \caption{Spatial topology}
        \label{fig:batch3}
    \end{subfigure}
    \caption{Illustration of various batch organization methods. We show the mapping results and the trajectories in a large-scale toy dataset\cite{ding2019deepmapping}. Using a spatial group would produce the best mapping results compared to other mapping methods. Other methods fail because the network wrongly registers frames from different areas together. Note that (b) is adopted in DeepMapping~\cite{ding2019deepmapping}.}
    \label{fig:batch_orgainize}
    \vspace{-3mm}
\end{figure} 

Due to the lack of loop closure in DeepMapping, extending the method to a large-scale environment is challenging. Though loop closure can be easily integrated into a SLAM system (\eg through bundle adjustment), it is hard to implement in deep-learning-based mapping methods because of its non-differentiability.

\textbf{Effect of different batch organization.} Dividing the whole trajectory into multiple mini-batches is inevitable. Then what is the correct way of such division? We tested several batch arrangements to split the dataset as shown in \cref{fig:batch_orgainize}. In \cref{fig:batch1}, a random organization of batches leads to an inferior mapping result because the network tends to pull random pairs of frames together and register them. In \cref{fig:batch2}, the batches are organized sequentially, as in DeepMapping. They can be registered well locally, but the global mapping quality is still inferior. Not surprisingly, in \cref{fig:batch3}, using spatially adjacent neighbors to form a batch would yield the best result. The reason is that M-Net would pull together the frames in a batch and register them together. Meanwhile, loop closure is incorporated into the training process because the loop is closed when all spatially adjacent frames are registered correctly in the loop.


\textbf{Loop-closure-based batch organization.} We construct a batch that contains spatially adjacent neighbors ($A$ and $N$s) using map topology from off-the-shelf place recognition algorithms. The map topology is a connected graph where each node represents a frame, and each edge connects two spatially adjacent nodes. To construct the batches, given an anchor frame $A$, we find the top $k$ closest frames connected to the anchor frame in map topology and organize them into one batch. By doing so, we can include loop closure information in the training process by letting M-Net register those spatially adjacent frames. This also makes it possible to construct the pairwise constraint in the local-to-global point consistency loss explained in \cref{sec:consistency}.

\subsection{Local-to-global point consistency loss}\label{sec:consistency}

\begin{figure}
    \centering
    \begin{subfigure}{0.19\textwidth}
        \includegraphics[width=\textwidth]{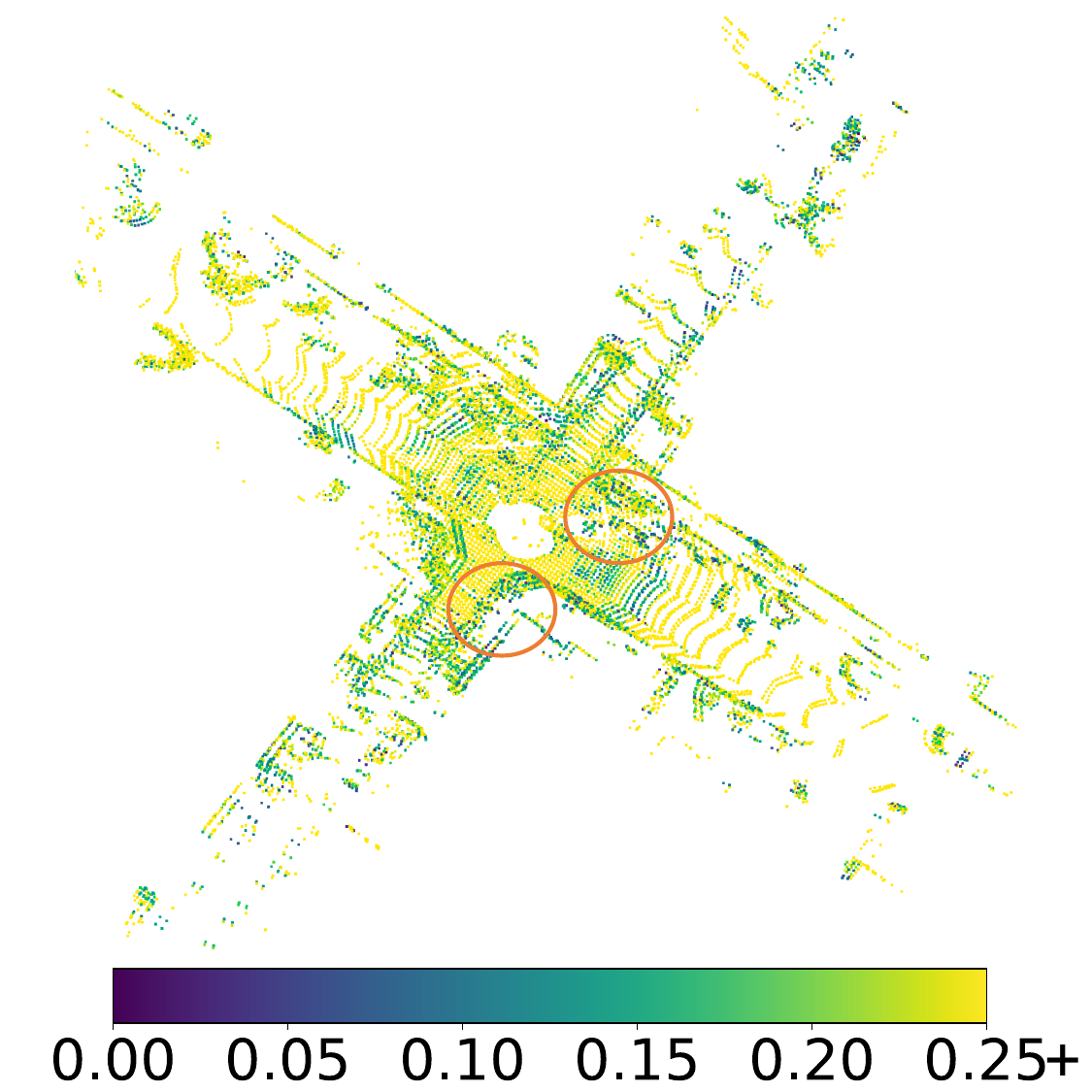}
        \caption{Color-coded distance (m)}
        \label{fig:color_dist}
    \end{subfigure}
    \begin{subfigure}{0.19\textwidth}
        \includegraphics[width=\textwidth]{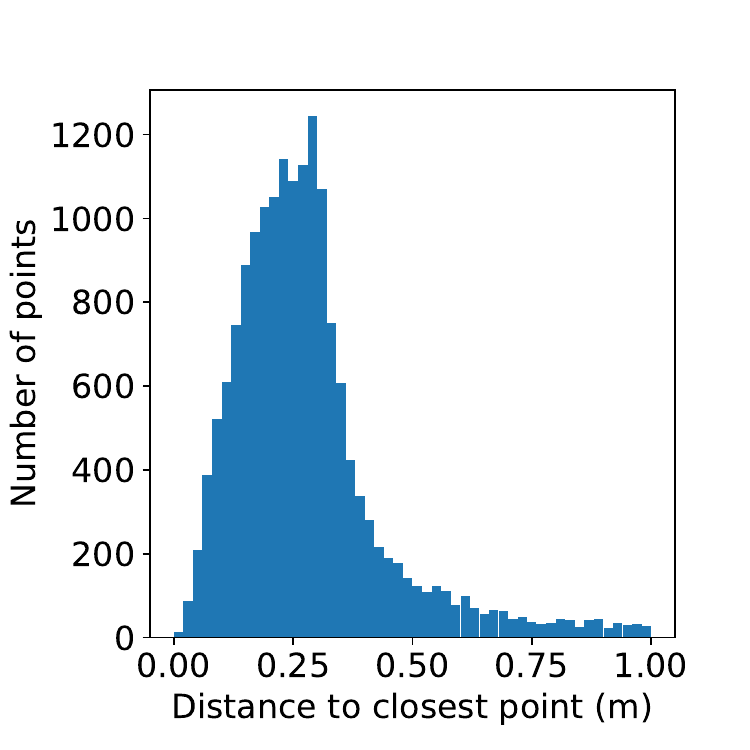}
        \caption{Histogram of distance}
        \label{fig:hist_dist}
    \end{subfigure}
    \vspace{-1mm}
    \caption{\textbf{Example of LiDAR point clouds lacking point-to-point correspondence}. We select two consecutive frames from the KITTI~\cite{geiger2013vision} dataset and register them with ground truth transformation. The distance between the two sensor poses is 1.09 m. For each point in the point cloud, we compute its distance to the closest point after registration. The color-coded distance is shown in (a) and the histogram in (b). The place where points have a distance smaller than 0.05 m is marked by red circles.}
    \label{fig:close_dist}
    \vspace{-3.5mm}
\end{figure}

\textbf{Slow convergence.} Even with the new batch organization, DeepMapping can still converge slowly, especially in large-scale scenes. Although initialization can give a "warm start" to the network, this information becomes decrepit as the training gets farther away from the initialization because it does not provide enough constraint on global pose estimation. Therefore, we want to incorporate the information from the initialization into the training process to provide a global inference cue for the networks. This can be done by constraining the pairwise relations of two adjacent point clouds to minimize their distance in the global frame. 

\textbf{Point correspondence.} However, point-level correspondence is needed to compute the distance between points. This correspondence is rare in large-scale and/or outdoor scenarios where mapping algorithms are often applied. As shown in \cref{fig:close_dist}, most points are about 0.25 m away from their closest points in the real-world dataset. Although alternative distance can be calculated by finding the closest points or abstractly described by hand-crafted descriptors~\cite{steder2010robust} or learning-based features~\cite{wang2019deep}, they can be inaccurate because point-to-point correspondence may not exist in the first place. When two point clouds are scanned at different sensor poses, it is very unlikely that the same point would exist in both point clouds due to the sparsity of LiDAR scanning. This makes the correspondence found by the methods above inaccurate. Thus, we ask the question: How can we constrain the pairwise relations of globally registered point clouds while not relying on inaccurate point-to-point correspondence?


\textbf{Distance metric.} We approach the question by considering the points in a single point cloud. When the point cloud $S$ is transformed by different transformation matrices $T$, the point-to-point correspondence is preserved because they are from the same local point cloud. After transformation, the distance between the two point clouds can be easily calculated by averaging the L2 distance between each corresponding point. The metric can be defined as a function of two transformations $T$, $T'$ and one single point cloud $S$:
\vspace{-1mm}
\begin{equation}\label{eq:distance}
    d(T,T',S)=\frac{1}{|S|}\sum_{s\in S}\lVert Ts-T's \rVert_2.
    \vspace{-1mm}
\end{equation}
Note that \cref{eq:distance} not only measures the distance between the two point clouds after transformation but also reflects the difference (inconsistency) between the two transformations. This is desired in our problem because a relatively accurate pairwise transformation is available from each neighbor frame to the anchor frame. We want to constrain the estimation of L-Net so that the pairwise relations are preserved in the global sensor poses. 

\begin{figure}
    \centering
    \begin{subfigure}{0.19\textwidth}
        \includegraphics[width=\textwidth]{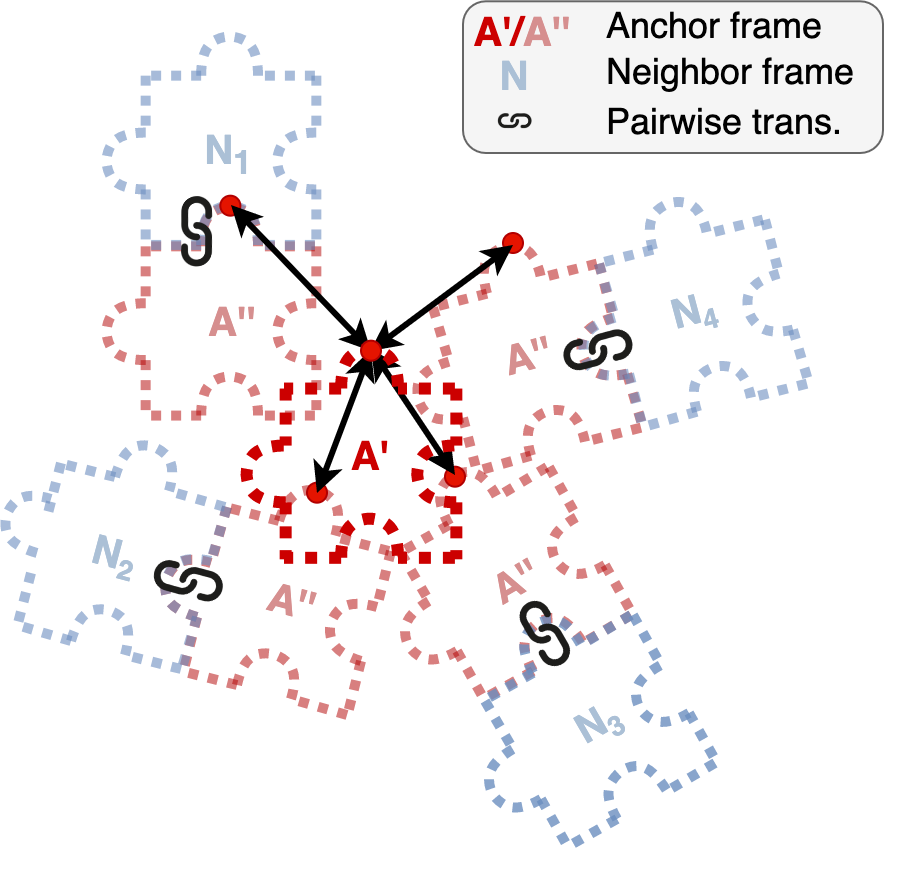}
        \caption{Poor global registration}
        \label{fig:consistancy1}
    \end{subfigure}
    \begin{subfigure}{0.19\textwidth}
        \includegraphics[width=\textwidth]{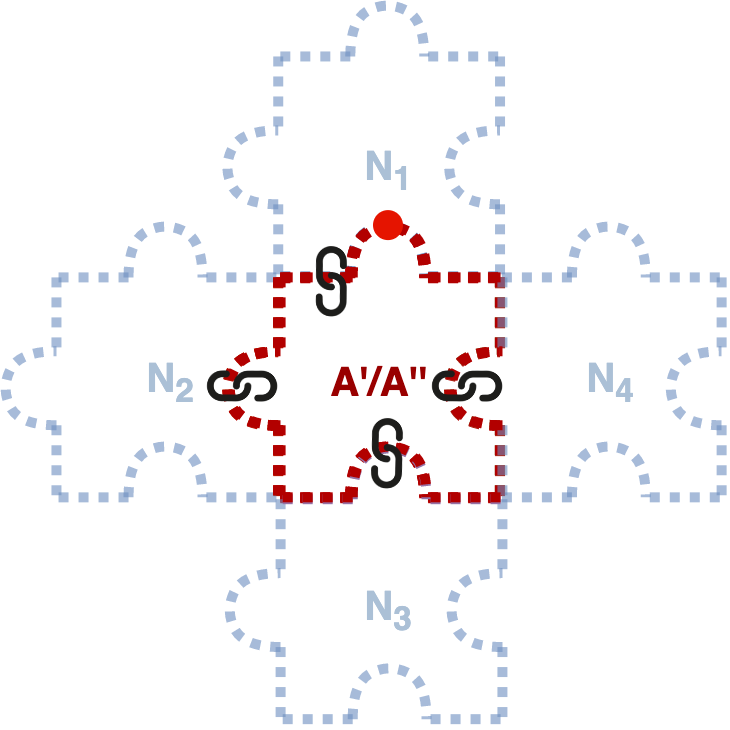}
        \caption{Perfect global registration}
        \label{fig:consistancy2}
    \end{subfigure}
    \vspace{-3pt}
    \caption{\textbf{Illustration of consistency loss.} \textcolor{red}{Red} puzzle represents different versions of anchor frames and \textcolor{cyan}{blue} puzzle represents the neighbor frames. The loss measures the euclidean distance of all the corresponding points of different versions of anchor frames that transformed from different poses, as indicated by the two-way arrows. \textcolor{red}{Red points} in the figure show one of the points for illustration.}
    \label{fig:consistency}
    \vspace{-4mm}
\end{figure}

\cref{fig:consistency} depicts the idea of this metric. Two versions of anchor frames are shown. $A'$ is transformed by the global pose estimated by L-Net. $A''$ is transformed by each neighbor frame's global pose and relative transformation. In \cref{fig:consistancy1}, the point clouds are poorly registered so the distances indicated by the arrows are large. In \cref{fig:consistancy2}, however, all corresponding points overlap so the distance is zero.

\textbf{Consistency loss.} Following this line of thought, we design the local-to-global consistency loss to measure the inconsistency between the local and global registrations. We denote the pairwise transformation between $S_i$ and $S_j$ as $T_i^j$, and the global pose of $S_i$ as $T_i^G$. Also, recall that neighbor $\mathcal{N}_i$ is defined as the indices of neighbor frames of $S_i$. We formulate the consistency loss as
\vspace{-2mm}
\begin{equation}\label{eq:consistency}
    \mathcal{L}_C= \frac{1}{K|\mathcal{N}_i|} \sum_{i=1}^{K} \sum_{j\in\mathcal{N}_i}d(T_j^GT_j^i, T_i^G,S_i).
\end{equation}
\vspace{-3mm}

It is worth noting that loss in \cref{eq:consistency} solves the two challenges mentioned above altogether. Global inference cue is provided while avoiding point-to-point correspondence. Because each $T_j^i$ is computed pairwisely and is considered more accurate than pairwise relations in the global estimation, by minimizing $\mathcal{L}_C$, the network can get more information from the initialization and have a faster convergence.

\section{Experiment}\label{sec:experiment}
\textbf{Dataset.} We use three datasets for a comprehensive evaluation: (1) KITTI dataset~\cite{geiger2013vision} for evaluations in outdoor scenarios, (2) NeBula odometry dataset~\cite{agha2021nebula} for evaluations in indoor environments where GPS signal is not available, and (3) NCLT dataset~\cite{carlevaris2016university} for evaluations in both scenarios. 

\textbf{Metrics.} For quantitative evaluations, we use the absolute trajectory error (ATE)~\cite{zhang2018tutorial} following DeepMapping~\cite{ding2019deepmapping}. For qualitative evaluations, we visualize both the registered map and the trajectory in the bird's eye view.

\textbf{Baselines.} We compare our method with baselines that fall into three different categories: (1) multiway registration~\cite{choi2015robust} is a \textit{multiple registration} method, (2) ICP~\cite{besl1992method}, DGR~\cite{choy2020deep}, HRegNet~\cite{lu2021hregnet}, GeoTransformer~\cite{qin2022geometric}, and KISS-ICP~\cite{vizzo2023kiss} are \textit{pairwise registration} algorithms that can run incrementally to obtain the global pose estimation, and (3) LeGO-LOAM~\cite{shan2018lego} is a \textit{SLAM} method. We also do ablation studies to compare the effectiveness of each proposed technique in \cref{sec:ablation}.

\textbf{Warm start}. Following DeepMapping~\cite{ding2019deepmapping}, this step has the same function as the front-end initialization in SfM before bundle adjustment (BA). In fact, DeepMapping2 can be seen as the deep learning version of BA for LiDAR mapping. Note that just like BA in SfM is affected by the initial solution's quality, DeepMapping2 is also affected by the ``warm start'' quality. Nonetheless, DeepMapping2 can be seamlessly integrated into almost any mapping front-end such as ICP, KISS-ICP, and Lego-LOAM described in the following sections, and always improves ATE without manual hyperparameter tuning.

\begin{figure*}[t]
    \centering
    \includegraphics[width=0.85\textwidth]{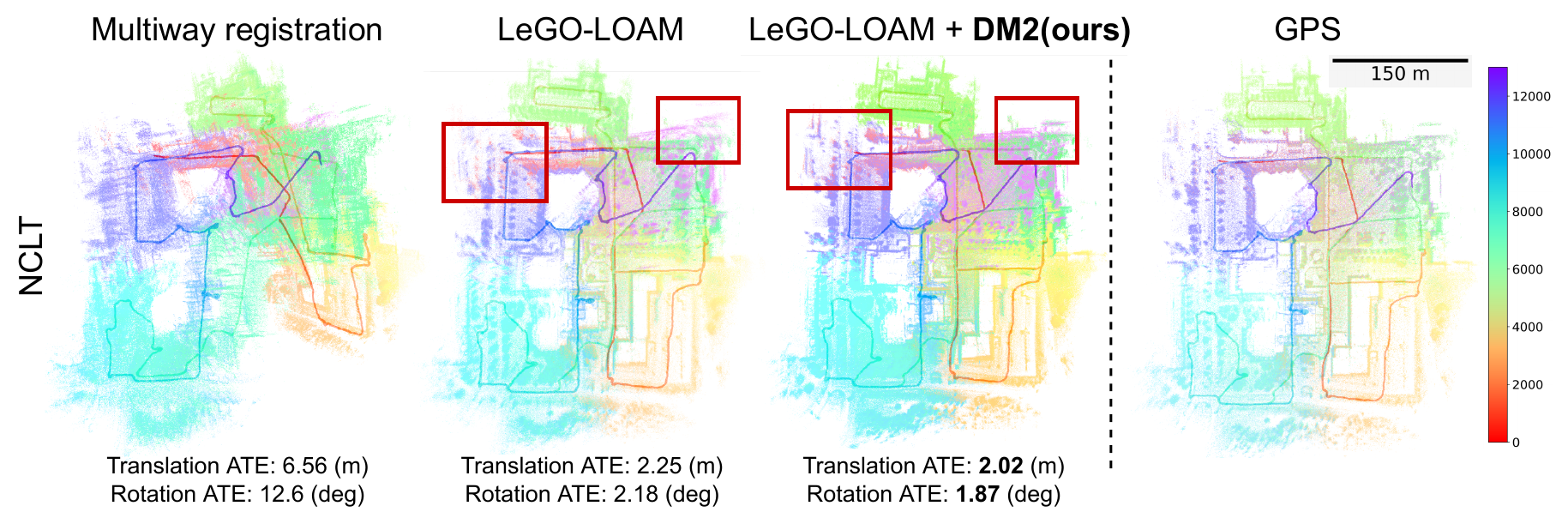}
    \caption{\textbf{Mapping results on NCLT dataset.} The red boxes indicate the places where our method has a better mapping quality than LeGO-LOAM. The ATEs are also listed. Other labels and legends follow \cref{fig:kitti_traj}. Best viewed in color.}
    \label{fig:nclt_traj}
    \vspace{-5pt}
\end{figure*}

\subsection{KITTI dataset}\label{sec:KITTI}


KITTI~\cite{geiger2013vision} is a widely-used authoritative benchmark to evaluate SLAM-based algorithms~\cite{chalvatzaras2022survey}. We employ the two most complex and challenging scenes from the dataset, where lots of places are revisited multiple times and the explored area is relatively large. Meanwhile, there are dynamic objects on the road, further increasing the difficulties.

\textbf{Quantitative comparisons.} From \cref{tab:kitti_ate}, we see that DeepMapping2 can perform well even with very trivial map initialization, like incremental ICP. There are significant improvements in the map quality when comparing the optimized map and the initialization, \eg, DeepMapping2 improves LeGO-LOAM by $16.6\%$ in terms of translation on drive\_0018. In general, our method is robust to initialization. No matter what category of methods is used to initialize the map, the performance of our method is consistent.

\textbf{Qualitative comparisons.}
As shown in ~\cref{fig:kitti_traj}, although multiway registration and DGR perform well at the start of the trajectory, the error is accumulated as the number of frames increases, leading to a noticeable drift. While the loop closure module in LeGO-LOAM is supposed to correct the drift, we can still see the errors on the trajectory, especially when the number of frames in drive\_0027 exceeds 4,000. On the other hand, the qualitative results produced by our approach do not have a noticeable difference from those provided by the GPS sensor.

\subsection{NCLT dataset}\label{sec:NCLT}
\begin{table}[t]
    \caption{\textbf{Quantitative result on the KITTI dataset.} The quality of the predicted trajectory is measured by absolute trajectory error (ATE) compared with the GPS trajectory. Results on two trajectories are listed. ``T-ATE" means translation ATE (m) ``R-ATE" means rotation ATE (deg). The last two rows are our results.}
    \label{tab:kitti_ate}
    \centering
    \resizebox{\columnwidth}{!}{%
    \begin{tabular}{@{}lcccc@{}}
        \toprule
        \multirow{2}{*}{\textbf{Method}} & \multicolumn{2}{c}{\textbf{Drive\_0018}} & \multicolumn{2}{c}{\textbf{Drive\_0027}} \\
        & \textbf{T-ATE (m)}$\downarrow$ & \textbf{R-ATE ($^{\circ}$)}$\downarrow$ & \textbf{T-ATE (m)}$\downarrow$ & \textbf{R-ATE ($^{\circ}$)}$\downarrow$ \\
        \midrule
        Incremental ICP & 4.38 & 4.61 & 3.53 & 2.67 \\
        Multiway~\cite{choi2015robust} & 2.24 & 1.75 & 4.70 & 5.93 \\
        DGR~\cite{choy2020deep} & 3.15 & 4.09 & 4.12 & 1.59 \\
        LeGO-LOAM~\cite{shan2018lego} & 1.90 & 1.36 & 2.96 & 2.36 \\
        HRegNet~\cite{lu2021hregnet} & 30.61 & 94.90 & 45.49 & 85.36 \\
        GeoTransformer~\cite{qin2022geometric} & 4.03 & 3.02 & 10.15 & 15.34 \\
        ICP+DM~\cite{ding2019deepmapping} & 3.42 & 1.66 & 3.39 & 2.70 \\
        KISS-ICP~\cite{vizzo2023kiss} & 2.10 & 0.68 & 6.25 & 1.21 \\
        \midrule
        ICP+\textbf{DM2} & 1.81 & 0.72 & \textbf{2.29} & 1.57 \\
        KISS-ICP+\textbf{DM2} & 1.78 & \textbf{0.68} & 2.30 & \textbf{1.17} \\
        LeGO-LOAM+\textbf{DM2} & \textbf{1.63} & 1.18 & 2.59 & 2.27 \\

        \bottomrule
    \end{tabular}
    }
\vspace{-12pt}
\end{table}

NCLT~\cite{carlevaris2016university} is a large-scale dataset collected at the University of Michigan’s North Campus. The point clouds are collected using a Velodyne HDL-32E 3D LiDAR mounted on a Segway robot. In total, the NCLT dataset has a robot trajectory with a length of 147.4 km and maintains 27 discrete mapping sessions over the year. Each mapping session includes both indoor and outdoor environments. We select an interval of the trajectory for better illustration.

\textbf{Qualitative results} As shown in \cref{fig:nclt_traj}, although LeGO-LOAM can produce a relatively satisfactory map, it still misaligns point clouds in some areas of the trajectory. In the red boxes in \cref{fig:nclt_traj}, the optimized map by \titlevariable~has better alignment than that from LeGO-LOAM, which is also proved by the ATEs reported under the maps.

\textbf{Quantitative results} The translation and rotation ATEs of different methods on NCLT are reported in \cref{tab:nclt_ate}. Incremental ICP does not have good map quality. Hence, when using it as initialization, our method does not have a better map than LeGO-LOAM, despite the fact that it reduces the errors by nearly one-half. Nevertheless, our method can still improve the map from LeGO-LOAM and have lower ATEs. 

It is worth noting that the slightly better map quality by \cite{shan2018lego} mainly results from the smaller area and more frames in the trajectories (shown by the scale in \cref{fig:nclt_traj}). However, this is not always expected in all implementations. Comparing the results in KITTI and NCLT, we find our method having a consistent improvement to the initialization regardless of the sampling density of the environment. 


\begin{table}[t]
\scriptsize
\renewcommand\tabcolsep{10pt}
    \caption{\textbf{Quantitative results on the NCLT dataset}. The notation is the same as in \cref{tab:kitti_ate}. The last two rows are our results.}
    \label{tab:nclt_ate}
    \centering
    \begin{tabular}{@{}lcc@{}}
        \toprule
        \textbf{Method} & \textbf{T-ATE (m)}$\downarrow$ & \textbf{R-ATE ($^{\circ}$)}$\downarrow$ \\
        \midrule
        Incremental ICP & 6.20 & 12.95 \\
        Multiway~\cite{choi2015robust} & 6.56	& 12.6 \\
        DGR~\cite{choy2020deep} & 8.89 & 42.9 \\
        LeGO-LOAM~\cite{shan2018lego} & 2.25 & 2.18 \\
        \midrule
        ICP+\textbf{DM2} & 3.73 & 6.27 \\
        LeGO-LOAM+\textbf{DM2} & \textbf{2.02} & \textbf{1.87} \\
        \bottomrule
    \end{tabular}
    \vspace{-8pt}
\end{table}

\subsection{NeBula dataset}\label{sec:Nebula}
\begin{figure*}[t]
    \centering
    \includegraphics[width=1\textwidth]{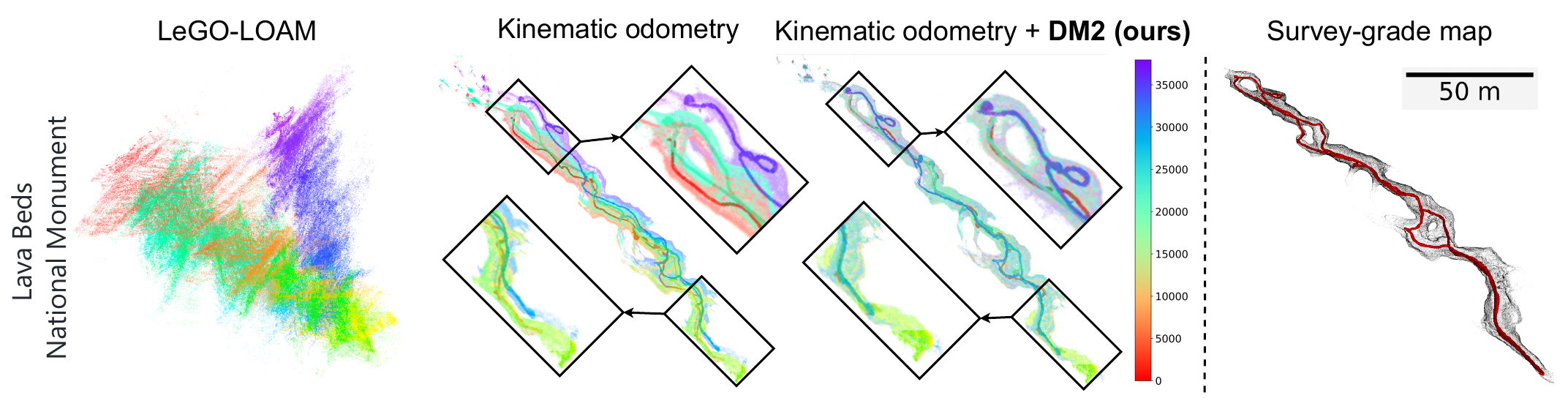}
    \caption{\textbf{Mapping result on NeBula dataset.} Several parts of the map are zoomed in for a clearer demonstration. As far as we know, we are the first to show this quality of mapping result of the trajectory in the Nebula dataset. Other labels and legends follow \cref{fig:kitti_traj}. }
    \label{fig:nebula_traj}
    \vspace{-8pt}
\end{figure*}

NeBula odometry\cite{agha2021nebula} dataset is provided by Team CoSTAR. The dataset is used to test the multi-robot system in real-world environments such as the DARPA Subterranean Challenge~\cite{rouvcek2019darpa}. In each scene of the dataset, a ground-truth survey-grade map is provided by DARPA and the trajectory is produced by running LOCUS 2.0~\cite{reinke2022locus}. Visual odometry and kinematic odometry are also included. We test different methods on the data collected in Lava Beds National Monument. The whole trajectory is $590.85$ meters long and contains 37,949 LiDAR scans in total. 

\textbf{Improvement to the odometry.}\label{sec:improve} 
Nebula is very challenging because it contains a long indoor trajectory without any marker for loop closure detection. We first run initialization on the dataset with incremental ICP and LeGO-LOAM. However, the mapping result is very poor (see \cref{fig:nebula_traj}) and cannot be used as a "warm start" for DeepMapping2. We find, however, the provided kinematic odometry  gives a good mapping result but can still be optimized. Hence, we run DeepMapping2 with the initialization from the kinematic odometry. As shown in \cref{fig:nebula_traj}, the optimized map corrects the large misalignments at the two ends of the tunnel. It is very accurate qualitatively compared to the survey-grade map.
This dataset does not provide the complete trajectory so we cannot do a quantitative analysis. 

\subsection{Ablation study}\label{sec:ablation}

We do experiments on the KITTI dataset to analyze how the proposed techniques influence the mapping quality. We compare the quantitative results when one or two of the components are missing from the pipeline.

\begin{figure}
    \centering
    \includegraphics[width=0.5\textwidth]{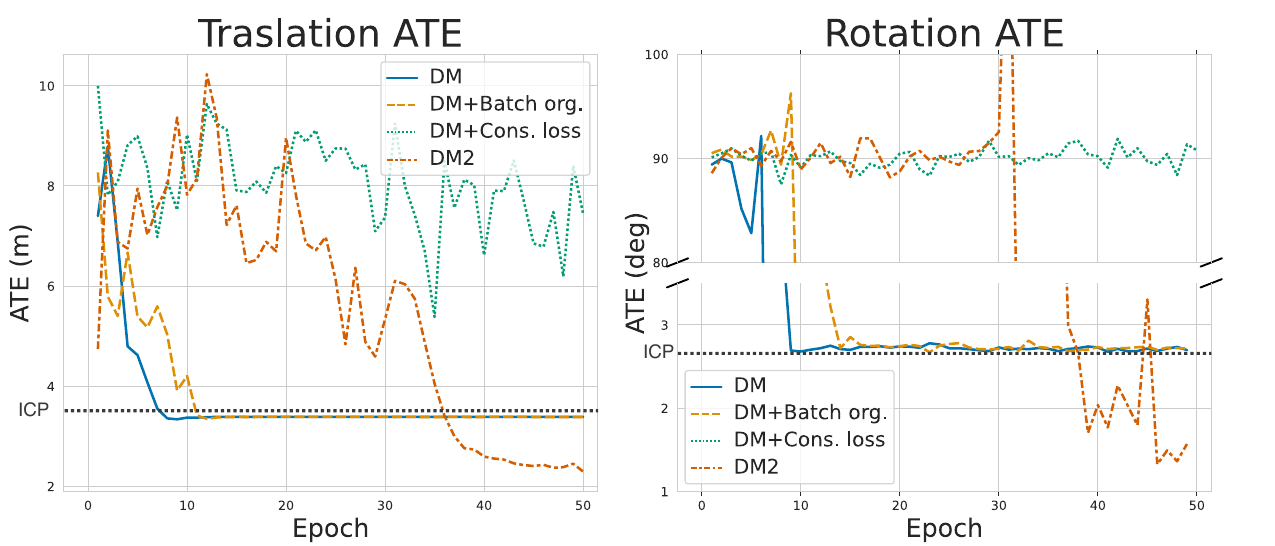}
    \caption{\textbf{ATE versus training epoch on drive\_0027 in KITTI.} The legend indicates the ATEs with different components in \cref{tab:kitti_ablation}.}
    \label{fig:training_curve}
    \vspace{-20pt}
\end{figure}


\textbf{Batch organization} As shown in \cref{tab:kitti_ablation}, batch organization plays an important role in mapping. The first version of DeepMapping (DM) constructs a map with batches organized sequentially. Drifts will occur when encountering large-scale scenarios with many revisits. On the other hand, our method can have a better mapping quality with batch organization due to implicit loop closure.

\textbf{Consistency loss.} Despite the benefits of local-to-global consistency loss mentioned in \cref{sec:consistency}, it requires accurate pairwise registrations to work. Row 1 and row 3 in \cref{tab:kitti_ablation} show that adding consistency loss alone to DeepMapping loss does not lead to a significant improvement. This is because the pairwise registration from off-the-shelf algorithms is prone to errors for non-adjacent point clouds. Comparing row 2 to row 5 in \cref{tab:kitti_ablation}, only when batch organization groups spatially adjacent frames together, can consistency loss work ideally and have the best mapping quality. It can also be seen from \cref{fig:training_curve} that consistency loss speed-ups the convergence.



\begin{table}
\scriptsize
    \caption{\textbf{Ablation study on the KITTI dataset.} The experiment is done by combining different components in the pipeline of \titlevariable. All the results reported are from the 50th epoch of the training. The method without DeepMapping loss fails to converge so its result is not reported.}
    \label{tab:kitti_ablation}
    \centering
    \resizebox{\columnwidth}{!}{%
    \begin{tabular}{@{}ccccc@{}}
        \toprule
        \multicolumn{3}{c}{\textbf{Components}} & \multirow{2}{*}{\textbf{T-ATE (m)}$\downarrow$} & \multirow{2}{*}{\textbf{R-ATE ($^{\circ}$)}$\downarrow$} \\
        DM loss & Batch org. & Con. loss \\
        \midrule
        \checkmark & & & 1.88 & 4.72 \\ 
        \checkmark & \checkmark & & 1.65 & 2.07\\ 
        \checkmark & & \checkmark & 1.88 & 4.70 \\ 
         & \checkmark & \checkmark & --- & --- \\
        \checkmark & \checkmark & \checkmark & \textbf{1.63} & \textbf{1.81} \\ 
        \bottomrule
    \end{tabular}
    }
    \vspace{-10pt}

\end{table}

\section{Conclusion}\label{sec:conclusion}

\textbf{{Limitation}.} \titlevariable~requires good loop closure to arrange batches. However, sometimes in large-scale environments, using a self-supervised loop closing method such as TF-VPR\cite{chen2022self} is time-comsuming. One alternative is to use pretrained PointNetVLAD\cite{uy2018pointnetvlad} with some geometric verification, such as ICP. Another option is to obtain loop closure from the pre-built map used in the warm-start step.

Also, although our convergence rate is improved by the \lgloss, the computation time of our method is still longer than other state-of-the-art methods like LeGO-LOAM. However, our framework allows \textit{GPU-based parallel optimization} for speed boosting via distributed data parallel training~\cite{li2020pytorch} as in the supplementary.

\textbf{{Summary}.} \titlevariable~ adds loop closure-based batch organization and self-supervised \lgloss~to DeepMapping, which can achieve excellent mapping performance in large-scale scenes. We believe our work can motivate more research in self-supervised LiDAR map optimization. The current pipeline can be used as a generic learning-based map optimization method in the back-end of almost any point cloud SLAM methods. Future works include adopting ideas like keyframe-based SLAM, and generalizing \titlevariable~to multi-agent SLAM.

\small{
\textbf{Acknowledgment}. This work is supported by NSF CMMI-1932187, CNS-2121391, EEC-2036870, and IIS-2238968.
}

{\small
\balance
\bibliographystyle{unsrt}
\bibliography{egbib}

\begin{thebibliography}{10}

\bibitem{ding2019deepmapping}
Li~Ding and Chen Feng.
\newblock Deepmapping: Unsupervised map estimation from multiple point clouds.
\newblock In {\em CVPR}, pages 8650--8659, 2019.

\bibitem{wolcott2014visual}
Ryan~W Wolcott and Ryan~M Eustice.
\newblock Visual localization within lidar maps for automated urban driving.
\newblock In {\em 2014 IEEE/RSJ International Conference on Intelligent Robots
  and Systems}, pages 176--183. IEEE, 2014.

\bibitem{lu2019l3}
Weixin Lu, Yao Zhou, Guowei Wan, Shenhua Hou, and Shiyu Song.
\newblock L3-net: Towards learning based lidar localization for autonomous
  driving.
\newblock In {\em CVPR}, pages 6389--6398, 2019.

\bibitem{palieri2020locus}
Matteo Palieri, Benjamin Morrell, Abhishek Thakur, Kamak Ebadi, Jeremy Nash,
  Arghya Chatterjee, Christoforos Kanellakis, Luca Carlone, Cataldo
  Guaragnella, and Ali-akbar Agha-Mohammadi.
\newblock Locus: A multi-sensor lidar-centric solution for high-precision
  odometry and 3d mapping in real-time.
\newblock {\em IEEE Robotics and Automation Letters}, 6(2):421--428, 2020.

\bibitem{taguchi2013point}
Yuichi Taguchi, Yong-Dian Jian, Srikumar Ramalingam, and Chen Feng.
\newblock Point-plane slam for hand-held 3d sensors.
\newblock In {\em 2013 IEEE international conference on robotics and
  automation}, pages 5182--5189. IEEE, 2013.

\bibitem{sarlin2022lamar}
Paul-Edouard Sarlin, Mihai Dusmanu, Johannes~L. Sch\"onberger, Pablo Speciale,
  Lukas Gruber, Viktor Larsson, Ondrej Miksik, and Marc Pollefeys.
\newblock {LaMAR}: {B}enchmarking {L}ocalization and {M}apping for {A}ugmented
  {R}eality.
\newblock In {\em ECCV}, 2022.

\bibitem{davison2007monoslam}
Andrew~J Davison, Ian~D Reid, Nicholas~D Molton, and Olivier Stasse.
\newblock Monoslam: Real-time single camera slam.
\newblock {\em PAMI}, 29(6):1052--1067, 2007.

\bibitem{klein2007parallel}
Georg Klein and David Murray.
\newblock Parallel tracking and mapping for small ar workspaces.
\newblock In {\em 2007 6th IEEE and ACM international symposium on mixed and
  augmented reality}, pages 225--234. IEEE, 2007.

\bibitem{izadi2011kinectfusion}
Shahram Izadi, David Kim, Otmar Hilliges, David Molyneaux, Richard Newcombe,
  Pushmeet Kohli, Jamie Shotton, Steve Hodges, Dustin Freeman, Andrew Davison,
  et~al.
\newblock Kinectfusion: real-time 3d reconstruction and interaction using a
  moving depth camera.
\newblock In {\em Proceedings of the 24th annual ACM symposium on User
  interface software and technology}, pages 559--568, 2011.

\bibitem{engel2014lsd}
Jakob Engel, Thomas Sch{\"o}ps, and Daniel Cremers.
\newblock Lsd-slam: Large-scale direct monocular slam.
\newblock In {\em ECCV}, pages 834--849. Springer, 2014.

\bibitem{labbe2019rtab}
Mathieu Labb{\'e} and Fran{\c{c}}ois Michaud.
\newblock Rtab-map as an open-source lidar and visual simultaneous localization
  and mapping library for large-scale and long-term online operation.
\newblock {\em Journal of Field Robotics}, 36(2):416--446, 2019.

\bibitem{campos2021orb}
Carlos Campos, Richard Elvira, Juan J~G{\'o}mez Rodr{\'\i}guez, Jos{\'e}~MM
  Montiel, and Juan~D Tard{\'o}s.
\newblock Orb-slam3: An accurate open-source library for visual,
  visual--inertial, and multimap slam.
\newblock {\em IEEE Transactions on Robotics}, 37(6):1874--1890, 2021.

\bibitem{zhou2021lidar}
Lipu Zhou, Daniel Koppel, and Michael Kaess.
\newblock Lidar slam with plane adjustment for indoor environment.
\newblock {\em IEEE Robotics and Automation Letters}, 6(4):7073--7080, 2021.

\bibitem{zhang2014loam}
Ji~Zhang and Sanjiv Singh.
\newblock Loam: Lidar odometry and mapping in real-time.
\newblock In {\em Robotics: Science and Systems}, volume~2, pages 1--9.
  Berkeley, CA, 2014.

\bibitem{shan2018lego}
Tixiao Shan and Brendan Englot.
\newblock Lego-loam: Lightweight and ground-optimized lidar odometry and
  mapping on variable terrain.
\newblock In {\em 2018 IEEE/RSJ International Conference on Intelligent Robots
  and Systems}, pages 4758--4765. IEEE, 2018.

\bibitem{shan2020lio}
Tixiao Shan, Brendan Englot, Drew Meyers, Wei Wang, Carlo Ratti, and Daniela
  Rus.
\newblock Lio-sam: Tightly-coupled lidar inertial odometry via smoothing and
  mapping.
\newblock In {\em 2020 IEEE/RSJ international conference on intelligent robots
  and systems (IROS)}, pages 5135--5142. IEEE, 2020.

\bibitem{ebadi2020lamp}
Kamak Ebadi, Yun Chang, Matteo Palieri, Alex Stephens, Alex Hatteland, Eric
  Heiden, Abhishek Thakur, Nobuhiro Funabiki, Benjamin Morrell, Sally Wood,
  et~al.
\newblock Lamp: Large-scale autonomous mapping and positioning for exploration
  of perceptually-degraded subterranean environments.
\newblock In {\em 2020 IEEE International Conference on Robotics and Automation
  (ICRA)}, pages 80--86. IEEE, 2020.

\bibitem{kaess2008isam}
Michael Kaess, Ananth Ranganathan, and Frank Dellaert.
\newblock isam: Incremental smoothing and mapping.
\newblock {\em IEEE Transactions on Robotics}, 24(6):1365--1378, 2008.

\bibitem{kummerle2011g}
Rainer K{\"u}mmerle, Giorgio Grisetti, Hauke Strasdat, Kurt Konolige, and
  Wolfram Burgard.
\newblock g$^{2}$o: A general framework for graph optimization.
\newblock In {\em 2011 IEEE International Conference on Robotics and
  Automation}, pages 3607--3613. IEEE, 2011.

\bibitem{yew20183dfeat}
Zi~Jian Yew and Gim~Hee Lee.
\newblock 3dfeat-net: Weakly supervised local 3d features for point cloud
  registration.
\newblock In {\em ECCV}, pages 607--623, 2018.

\bibitem{shi2021keypoint}
Chenghao Shi, Xieyuanli Chen, Kaihong Huang, Junhao Xiao, Huimin Lu, and Cyrill
  Stachniss.
\newblock Keypoint matching for point cloud registration using multiplex
  dynamic graph attention networks.
\newblock {\em IEEE Robotics and Automation Letters}, 6(4):8221--8228, 2021.

\bibitem{li2019net}
Qing Li, Shaoyang Chen, Cheng Wang, Xin Li, Chenglu Wen, Ming Cheng, and
  Jonathan Li.
\newblock Lo-net: Deep real-time lidar odometry.
\newblock In {\em CVPR}, pages 8473--8482, 2019.

\bibitem{chen2021overlapnet}
Xieyuanli Chen, Thomas L{\"a}be, Andres Milioto, Timo R{\"o}hling, Olga
  Vysotska, Alexandre Haag, Jens Behley, and Cyrill Stachniss.
\newblock Overlapnet: Loop closing for lidar-based slam.
\newblock {\em arXiv preprint arXiv:2105.11344}, 2021.

\bibitem{kendall2015posenet}
Alex Kendall, Matthew Grimes, and Roberto Cipolla.
\newblock Posenet: A convolutional network for real-time 6-dof camera
  relocalization.
\newblock In {\em ICCV}, pages 2938--2946, 2015.

\bibitem{yang2015go}
Jiaolong Yang, Hongdong Li, Dylan Campbell, and Yunde Jia.
\newblock Go-icp: A globally optimal solution to 3d icp point-set registration.
\newblock {\em PAMI}, 38(11):2241--2254, 2015.

\bibitem{singandhupe2021registration}
Ashutosh Singandhupe, Hung La, Trung~Dung Ngo, and Van Ho.
\newblock Registration of 3d point sets using correntropy similarity matrix.
\newblock {\em arXiv preprint arXiv:2107.09725}, 2021.

\bibitem{wang2019deep}
Yue Wang and Justin~M Solomon.
\newblock Deep closest point: Learning representations for point cloud
  registration.
\newblock In {\em CVPR}, pages 3523--3532, 2019.

\bibitem{choy2020deep}
Christopher Choy, Wei Dong, and Vladlen Koltun.
\newblock Deep global registration.
\newblock In {\em CVPR}, 2020.

\bibitem{geiger2013vision}
Andreas Geiger, Philip Lenz, Christoph Stiller, and Raquel Urtasun.
\newblock Vision meets robotics: The kitti dataset.
\newblock {\em The International Journal of Robotics Research},
  32(11):1231--1237, 2013.

\bibitem{carlevaris2016university}
Nicholas Carlevaris-Bianco, Arash~K Ushani, and Ryan~M Eustice.
\newblock University of michigan north campus long-term vision and lidar
  dataset.
\newblock {\em The International Journal of Robotics Research},
  35(9):1023--1035, 2016.

\bibitem{agha2021nebula}
Ali Agha, Kyohei Otsu, Benjamin Morrell, David~D Fan, Rohan Thakker, Angel
  Santamaria-Navarro, Sung-Kyun Kim, Amanda Bouman, Xianmei Lei, Jeffrey
  Edlund, et~al.
\newblock Nebula: Quest for robotic autonomy in challenging environments; team
  costar at the darpa subterranean challenge.
\newblock {\em arXiv preprint arXiv:2103.11470}, 2021.

\bibitem{besl1992method}
Paul~J Besl and Neil~D McKay.
\newblock Method for registration of 3-d shapes.
\newblock In {\em Sensor fusion IV: control paradigms and data structures},
  volume 1611, pages 586--606. Spie, 1992.

\bibitem{maron2016point}
Haggai Maron, Nadav Dym, Itay Kezurer, Shahar Kovalsky, and Yaron Lipman.
\newblock Point registration via efficient convex relaxation.
\newblock {\em ACM Transactions on Graphics (TOG)}, 35(4):1--12, 2016.

\bibitem{theiler2015globally}
Pascal~Willy Theiler, Jan~Dirk Wegner, and Konrad Schindler.
\newblock Globally consistent registration of terrestrial laser scans via graph
  optimization.
\newblock {\em ISPRS journal of photogrammetry and remote sensing},
  109:126--138, 2015.

\bibitem{evangelidis2014generative}
Georgios~D Evangelidis, Dionyssos Kounades-Bastian, Radu Horaud, and
  Emmanouil~Z Psarakis.
\newblock A generative model for the joint registration of multiple point sets.
\newblock In {\em ECCV}, pages 109--122. Springer, 2014.

\bibitem{cui2015global}
Zhaopeng Cui and Ping Tan.
\newblock Global structure-from-motion by similarity averaging.
\newblock In {\em CVPR}, pages 864--872, 2015.

\bibitem{zhu2018very}
Siyu Zhu, Runze Zhang, Lei Zhou, Tianwei Shen, Tian Fang, Ping Tan, and Long
  Quan.
\newblock Very large-scale global sfm by distributed motion averaging.
\newblock In {\em CVPR}, pages 4568--4577, 2018.

\bibitem{choi2015robust}
Sungjoon Choi, Qian-Yi Zhou, and Vladlen Koltun.
\newblock Robust reconstruction of indoor scenes.
\newblock In {\em CVPR}, pages 5556--5565, 2015.

\bibitem{gojcic2020learning}
Zan Gojcic, Caifa Zhou, Jan~D Wegner, Leonidas~J Guibas, and Tolga Birdal.
\newblock Learning multiview 3d point cloud registration.
\newblock In {\em CVPR}, pages 1759--1769, 2020.

\bibitem{jegou2010aggregating}
Herv{\'e} J{\'e}gou, Matthijs Douze, Cordelia Schmid, and Patrick P{\'e}rez.
\newblock Aggregating local descriptors into a compact image representation.
\newblock In {\em CVPR}, pages 3304--3311. IEEE, 2010.

\bibitem{uy2018pointnetvlad}
Mikaela~Angelina Uy and Gim~Hee Lee.
\newblock Pointnetvlad: Deep point cloud based retrieval for large-scale place
  recognition.
\newblock In {\em CVPR}, pages 4470--4479, 2018.

\bibitem{qi2017pointnet}
Charles~R Qi, Hao Su, Kaichun Mo, and Leonidas~J Guibas.
\newblock Pointnet: Deep learning on point sets for 3d classification and
  segmentation.
\newblock In {\em CVPR}, pages 652--660, 2017.

\bibitem{arandjelovic2016netvlad}
Relja Arandjelovic, Petr Gronat, Akihiko Torii, Tomas Pajdla, and Josef Sivic.
\newblock Netvlad: Cnn architecture for weakly supervised place recognition.
\newblock In {\em CVPR}, pages 5297--5307, 2016.

\bibitem{chen2022self}
Chao Chen, Xinhao Liu, Xuchu Xu, Yiming Li, Li~Ding, Ruoyu Wang, and Chen Feng.
\newblock Self-supervised visual place recognition by mining temporal and
  feature neighborhoods.
\newblock {\em arXiv preprint arXiv:2208.09315}, 2022.

\bibitem{mur2015orb}
Raul Mur-Artal, Jose Maria~Martinez Montiel, and Juan~D Tardos.
\newblock Orb-slam: a versatile and accurate monocular slam system.
\newblock {\em IEEE transactions on robotics}, 31(5):1147--1163, 2015.

\bibitem{mur2017orb}
Raul Mur-Artal and Juan~D Tard{\'o}s.
\newblock Orb-slam2: An open-source slam system for monocular, stereo, and
  rgb-d cameras.
\newblock {\em IEEE transactions on robotics}, 33(5):1255--1262, 2017.

\bibitem{li2020deepslam}
Ruihao Li, Sen Wang, and Dongbing Gu.
\newblock Deepslam: A robust monocular slam system with unsupervised deep
  learning.
\newblock {\em IEEE Transactions on Industrial Electronics}, 68(4):3577--3587,
  2020.

\bibitem{bosse2009keypoint}
Michael Bosse and Robert Zlot.
\newblock Keypoint design and evaluation for place recognition in 2d lidar
  maps.
\newblock {\em Robotics and Autonomous Systems}, 57(12):1211--1224, 2009.

\bibitem{zlot2014efficient}
Robert Zlot and Michael Bosse.
\newblock Efficient large-scale 3d mobile mapping and surface reconstruction of
  an underground mine.
\newblock In {\em Field and service robotics}, pages 479--493. Springer, 2014.

\bibitem{steder2010robust}
Bastian Steder, Giorgio Grisetti, and Wolfram Burgard.
\newblock Robust place recognition for 3d range data based on point features.
\newblock In {\em 2010 IEEE International Conference on Robotics and
  Automation}, pages 1400--1405. IEEE, 2010.

\bibitem{zhang2018tutorial}
Zichao Zhang and Davide Scaramuzza.
\newblock A tutorial on quantitative trajectory evaluation for visual
  (-inertial) odometry.
\newblock In {\em 2018 IEEE/RSJ International Conference on Intelligent Robots
  and Systems}, pages 7244--7251. IEEE, 2018.

\bibitem{lu2021hregnet}
Fan Lu, Guang Chen, Yinlong Liu, Lijun Zhang, Sanqing Qu, Shu Liu, and Rongqi
  Gu.
\newblock Hregnet: A hierarchical network for large-scale outdoor lidar point
  cloud registration.
\newblock In {\em Proceedings of the IEEE/CVF International Conference on
  Computer Vision}, pages 16014--16023, 2021.

\bibitem{qin2022geometric}
Zheng Qin, Hao Yu, Changjian Wang, Yulan Guo, Yuxing Peng, and Kai Xu.
\newblock Geometric transformer for fast and robust point cloud registration.
\newblock In {\em CVPR}, pages 11143--11152, 2022.

\bibitem{vizzo2023kiss}
Ignacio Vizzo, Tiziano Guadagnino, Benedikt Mersch, Louis Wiesmann, Jens
  Behley, and Cyrill Stachniss.
\newblock Kiss-icp: In defense of point-to-point icp simple, accurate, and
  robust registration if done the right way.
\newblock {\em IEEE Robotics and Automation Letters}, 2023.

\bibitem{chalvatzaras2022survey}
Athanasios Chalvatzaras, Ioannis Pratikakis, and Angelos~A Amanatiadis.
\newblock A survey on map-based localization techniques for autonomous
  vehicles.
\newblock {\em IEEE Transactions on Intelligent Vehicles}, 2022.

\bibitem{rouvcek2019darpa}
Tom{\'a}{\v{s}} Rou{\v{c}}ek, Martin Pecka, Petr {\v{C}}{\'\i}{\v{z}}ek,
  Tom{\'a}{\v{s}} Pet{\v{r}}{\'\i}{\v{c}}ek, Jan Bayer, Vojt{\v{e}}ch
  {\v{S}}alansk{\`y}, Daniel He{\v{r}}t, Mat{\v{e}}j Petrl{\'\i}k,
  Tom{\'a}{\v{s}} B{\'a}{\v{c}}a, Voj{\v{e}}ch Spurn{\`y}, et~al.
\newblock Darpa subterranean challenge: Multi-robotic exploration of
  underground environments.
\newblock In {\em International Conference on Modelling and Simulation for
  Autonomous Systems}, pages 274--290. Springer, 2019.

\bibitem{reinke2022locus}
Andrzej Reinke, Matteo Palieri, Benjamin Morrell, Yun Chang, Kamak Ebadi, Luca
  Carlone, and Ali-Akbar Agha-Mohammadi.
\newblock Locus 2.0: Robust and computationally efficient lidar odometry for
  real-time 3d mapping.
\newblock {\em IEEE Robotics and Automation Letters}, 2022.

\bibitem{li2020pytorch}
Shen Li, Yanli Zhao, Rohan Varma, Omkar Salpekar, Pieter Noordhuis, Teng Li,
  Adam Paszke, Jeff Smith, Brian Vaughan, Pritam Damania, et~al.
\newblock Pytorch distributed: Experiences on accelerating data parallel
  training.
\newblock {\em arXiv preprint arXiv:2006.15704}, 2020.

\end{thebibliography}
}

\clearpage
\section*{Appendix}
\renewcommand{\thesection}{\Alph{section}}
\renewcommand{\thefigure}{\Roman{figure}}
\renewcommand{\thetable}{\Roman{table}}

\setcounter{section}{0}
\setcounter{figure}{0}
\setcounter{table}{0}

We provide in this supplementary more ablation studies, method analysis, and additional visualizations that could not fit in the paper. In particular, we include (1) the accuracy of various place recognition algorithms, (2) analysis of the computation time of our method and all baselines, (3) more visualizations including heat map and clearer mapping result, (4) and (5) video visualizations of trajectory estimation throughout the training processes.

\section{Robustness on map topology}

As mentioned in \cref{sec:pipeline}, the map topology used for organizing training batches can be obtained by any off-the-shelf algorithms. All results reported in \cref{sec:experiment} are based on TF-VPR~\cite{chen2022self}, which is a self-supervised place recognition algorithm. In this supplementary, we compare the mapping accuracy of DeepMapping2 when different approaches are used to provide map topology.

We compute the map topology from a pre-trained PointNetVLAD~\cite{uy2018pointnetvlad} model and GPS. The mapping results are shown in \cref{tab:robustness}. For drive\_0018, there is no significant difference between the mapping results from TF-VPR and PointNetVLAD. Also, using GPS provides a marginally better mapping result, but this is expected given that GPS is used as the ground truth. Due to PointNetVLAD's low-quality map topology, PointNetVLAD for drive\_0027 does not produce a good mapping result. In summary, \textbf{our method is robust regardless of the map topology used}, as long as it is relatively accurate to reflect the adjacency relationships in the environment. 

\begin{table}[ht]
    \vspace{-3mm}
    \caption{\textbf{Robustness on map topology.} The table lists the mapping result of DeepMapping2 when running with the map topology attained by different place recognition methods. GPS theoretically provides the most ideal map topology.} 
    \label{tab:robustness}
    \centering
    \resizebox{\columnwidth}{!}{%
    \begin{tabular}{@{}lcccc@{}}
        \toprule
        \multirow{2}{*}{\textbf{PR algo.}} & \multicolumn{2}{c}{\textbf{Drive\_0018}} & \multicolumn{2}{c}{\textbf{Drive\_0027}} \\
        & \textbf{T-ATE (m)}$\downarrow$ & \textbf{R-ATE ($^{\circ}$)}$\downarrow$ & \textbf{T-ATE (m)}$\downarrow$ & \textbf{R-ATE ($^{\circ}$)}$\downarrow$ \\
        \midrule
        TF-VPR~\cite{chen2022self} & 1.81 & 0.72 & 2.29 & 1.57 \\
        PointNetVLAD~\cite{uy2018pointnetvlad} & 1.82 & 0.80 & 4.79 & 7.80\\
        GPS (ground truth) & 1.62 & 0.62 & 2.07 & 1.42 \\
        \bottomrule
    \end{tabular}
    }
    \vspace{-3mm}
\end{table}

\section{Time analysis}

\begin{table}[ht]
    \caption{\textbf{Computation time of different methods}. The three baseline methods are run on CPU. DM2 is run on RTX3090 GPU. Note that there is no NVLink when 2 GPUs are used.}
    \label{tab:time}
    \vspace{-2mm}
    \centering
    \resizebox{0.8\linewidth}{!}{
    \begin{tabular}{@{}lcc@{}}
        \toprule
        \multirow{2}{*}{\textbf{Method}} & \multicolumn{2}{c}{\textbf{Time consumption (s)}} \\
        & \textbf{Drive\_0018} & \textbf{Drive\_0027} \\
        \midrule
        Multiway~\cite{choi2015robust} & 113 & 141 \\
        DGR~\cite{choy2020deep} (on CPU) & 70200 & 108522 \\
        LeGO-LOAM~\cite{shan2018lego} & 287 & 470 \\
        \midrule
        \textbf{DM2} (1 GPU) & 21600 & 29900 \\
        \textbf{DM2} (2 GPUs) & 12085 & 18587 \\
        \bottomrule
    \end{tabular}
    }
    \vspace{-2mm}
\end{table}

We compare the computation time of different methods on two trajectories from the KITTI~\cite{geiger2013vision} dataset. Due to the nature of training-as-optimization, as we mentioned in ~\cref{sec:conclusion}, our method takes longer to compute than some of the baselines. When we apply DeepMapping2 to larger datasets, we can apply parallel training to reduce computation time. In  \cref{tab:time}, it shows that if two GPUs are used for training, the training time is almost decreased by half. It is worth noting that our hardware has no NVLink, which is frequently used in distributed multi-GPU systems to speed up data transmission among GPUs. As a result, the theory and the actual scalability should be very similar. When given sufficient computational resources, the time needed by our method can be significantly reduced, \ie, the time required is expected to be \textbf{inversely proportional} to the number of GPUs used. Thus, DeepMapping2 should support a multi-agent setup where the point clouds are scanned by multiple agents and do not have a sequential order.

\section{More experiments}
\begin{figure*}[ht]
    \centering
        \begin{subfigure}{0.23\textwidth}
        \includegraphics[trim={0.5cm 0.5cm 11cm 0.5cm},clip,width=1\textwidth]{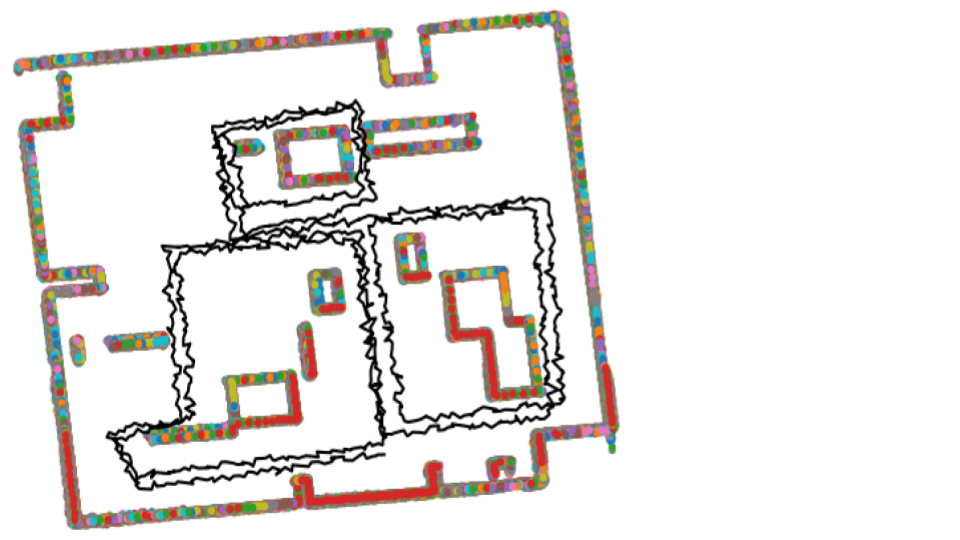}
        \caption{Scene1 (1024 frames)}
        \label{fig:dm-1024-1}
    \end{subfigure}
    \begin{subfigure}{0.23\textwidth}
        \includegraphics[trim={0.5cm 0.5cm 11cm 0.5cm},clip,width=1\textwidth]{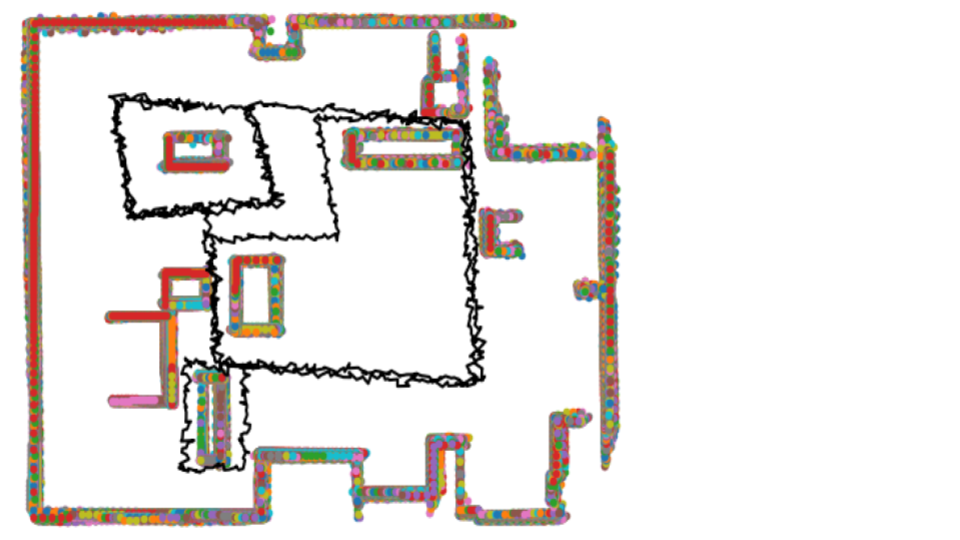}
        \caption{Scene2 (1024 frames)}
        \label{fig:dm-1024-2}
    \end{subfigure}
        \begin{subfigure}{0.23\textwidth}
        \includegraphics[trim={0.5cm 0.5cm 12cm 0.5cm},clip,width=1\textwidth]{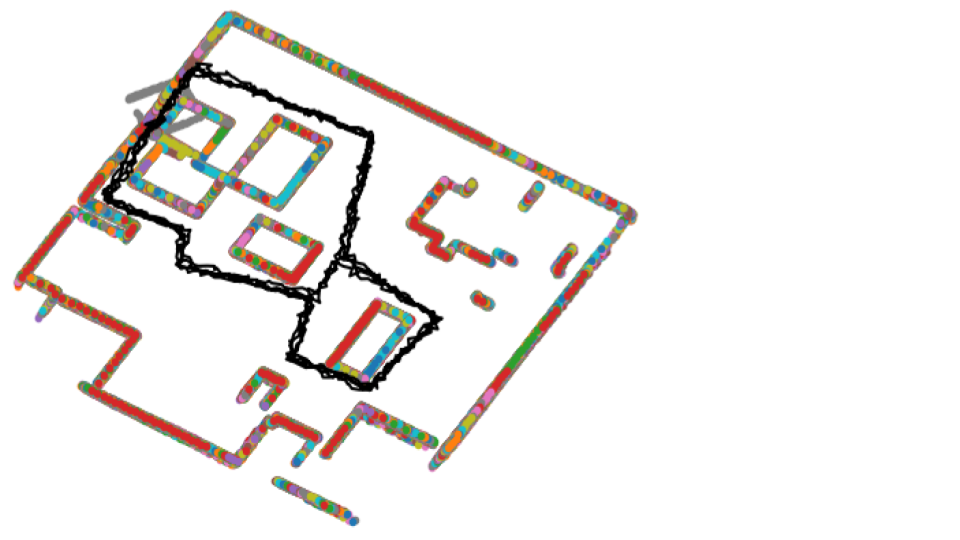}
        \caption{Scene3 (1024 frames)}
    \label{fig:dm-1024-3}
    \end{subfigure}
    \begin{subfigure}{0.23\textwidth}
        \includegraphics[trim={0.5cm 0.5cm 10cm 0.5cm},clip,width=1\textwidth]{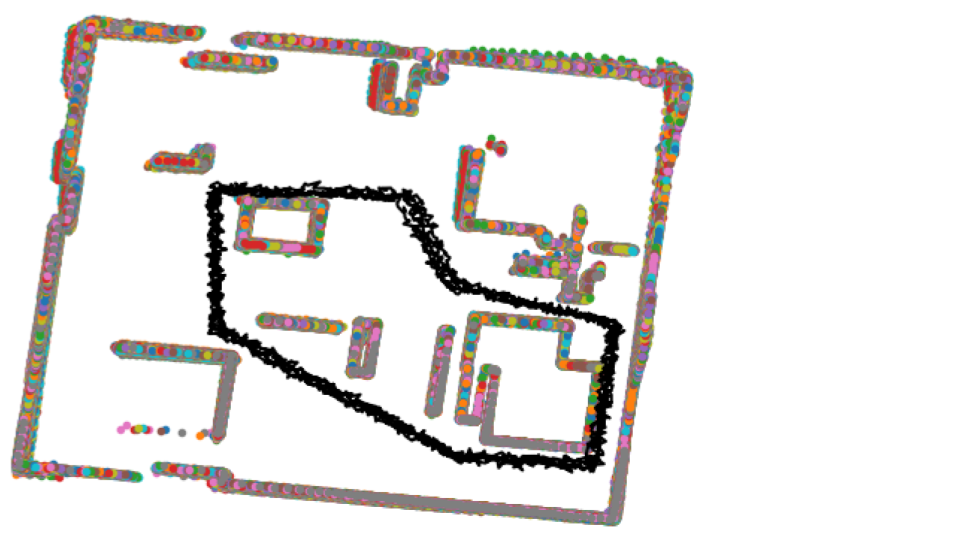}
        \caption{Scene4 (2048 frames)}
        \label{fig:dm-2048}
    \end{subfigure}
    \caption{\textbf{Mapping and trajectory plot on Simulated point cloud dataset~\cite{ding2019deepmapping}} We include five mapping results including (a)(b)(c) DeepMapping2 mapping results on three different trajectories with 1024 frames (d) DeepMapping2 mapping results on a trajectory with 2048 frames.}
    \label{fig:dm-point}
\end{figure*}

We conduct more tests on the simulated dataset~\cite{ding2019deepmapping}. The original DeepMapping pipeline fails on the large-scale dataset, because the drift cannot be correct as described in~\cref{sec:overview}. DeepMapping2 successfully estimates multiple trajectories with different numbers of frames on the simulated point dataset as visualized in \cref{fig:dm-point} 

We also conduct more experiments on other KITTI sequences~\cite{geiger2013vision}. The detailed result is shown in~\cref{tab:kitti_traj}.

\begin{table}[ht] 
\scriptsize
\renewcommand\tabcolsep{10pt}
    \caption{\textbf{Additional results on the KITTI dataset.}}
    \label{tab:kitti_traj}
    \vspace{-2mm}
    \centering
    \resizebox{\linewidth}{!}{%
    \begin{tabular}{@{}lcccc@{}}
        \toprule
        \multirow{2}{*}{\textbf{More sequence}} & \multicolumn{2}{c}{\textbf{Sequence 02}} & \multicolumn{2}{c}{\textbf{Sequence 08}} \\
        & \textbf{T-ATE (m)}$\downarrow$ & \textbf{R-ATE ($^{\circ}$)}$\downarrow$ & \textbf{T-ATE (m)}$\downarrow$ & \textbf{R-ATE ($^{\circ}$)}$\downarrow$ \\
        \midrule
        Incremental ICP & 8.06 & 4.57 & 4.38 & 4.46 \\
        Multiway Registration & 4.96 & 3.37 & 2.40 & 0.90 \\
        \midrule
        ICP+\textbf{DM2} & \textbf{2.56} & \textbf{1.31} & \textbf{1.85} & \textbf{0.81} \\
        \bottomrule
    \end{tabular}
    }
\vspace{-5mm}
\end{table}

\begin{figure}[ht]
    \centering
    \includegraphics[width=0.4\textwidth]{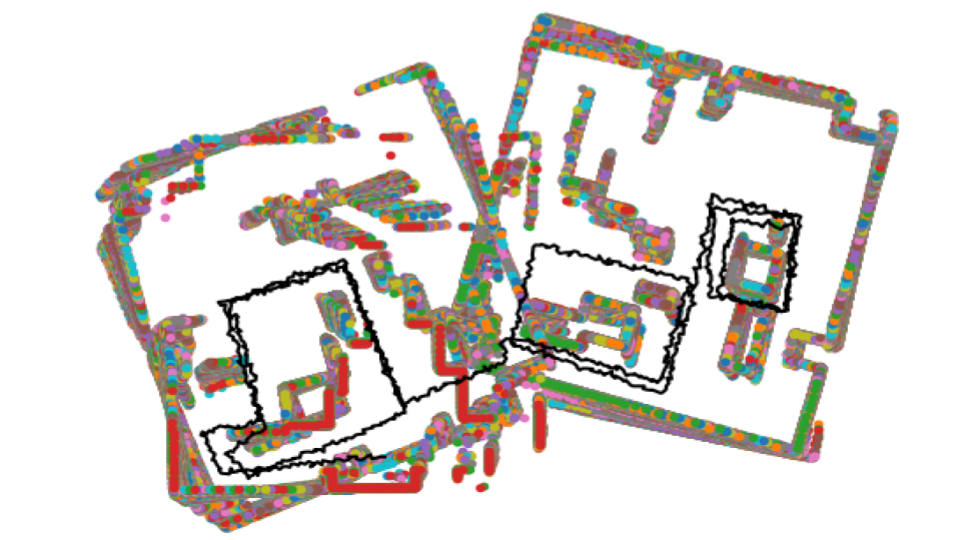}
    \caption{\textbf{Original DeepMapping results on simulated point cloud dataset.} The black line represents the trajectory, while the color block represents the occupancy grid.}
    \label{fig:dm-origin}
\end{figure}

\section{More visualization}
\begin{figure*}[ht]
    \centering
    \includegraphics[width=1\textwidth]{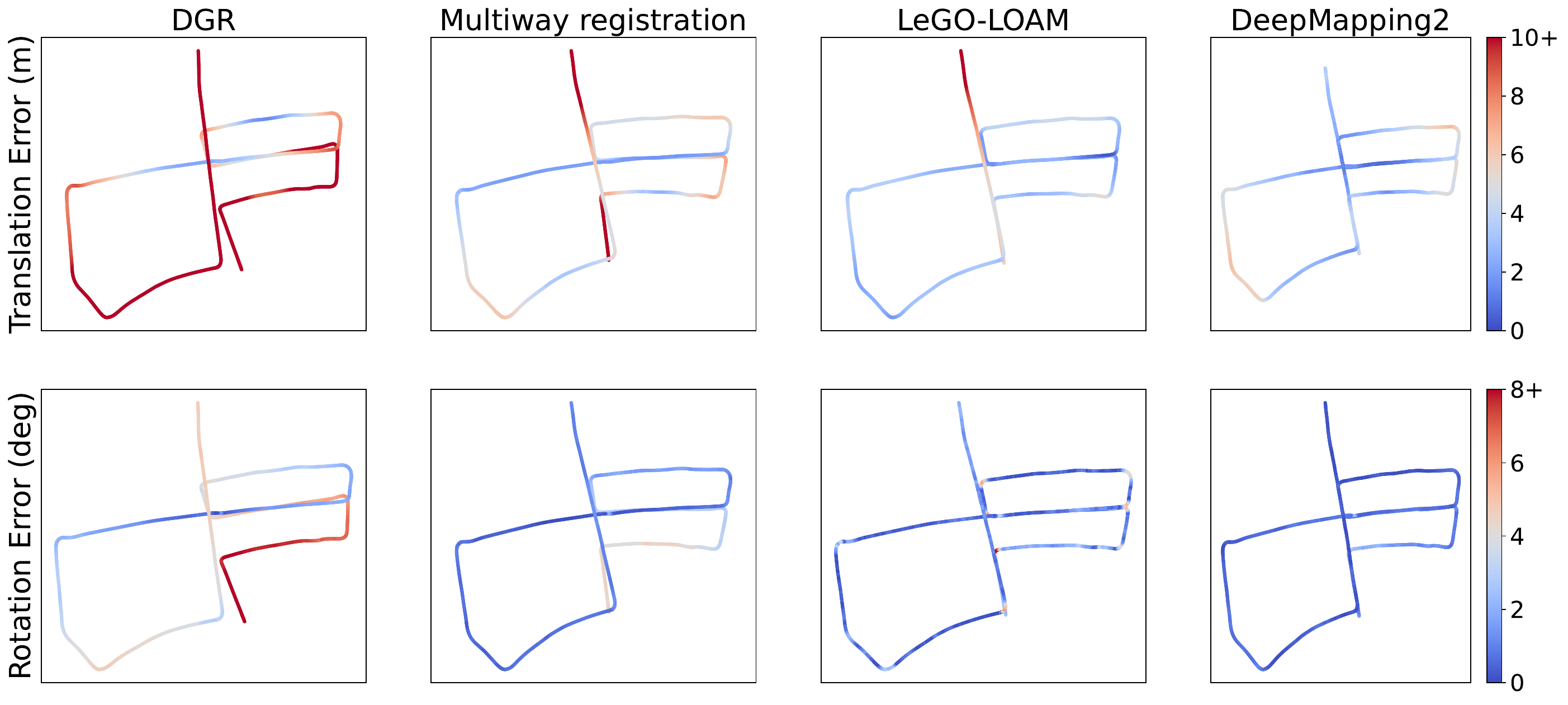}
    \caption{\textbf{KITTI drive\_0018.} Heat map visualization of both translation and rotation ATE for each frame in the dataset. Note that the color bar is clipped for better visualization. Best viewed in color.}
    \label{fig:heat_kitti18}
\end{figure*}

\begin{figure*}[ht]
    \centering
    \includegraphics[width=1\textwidth]{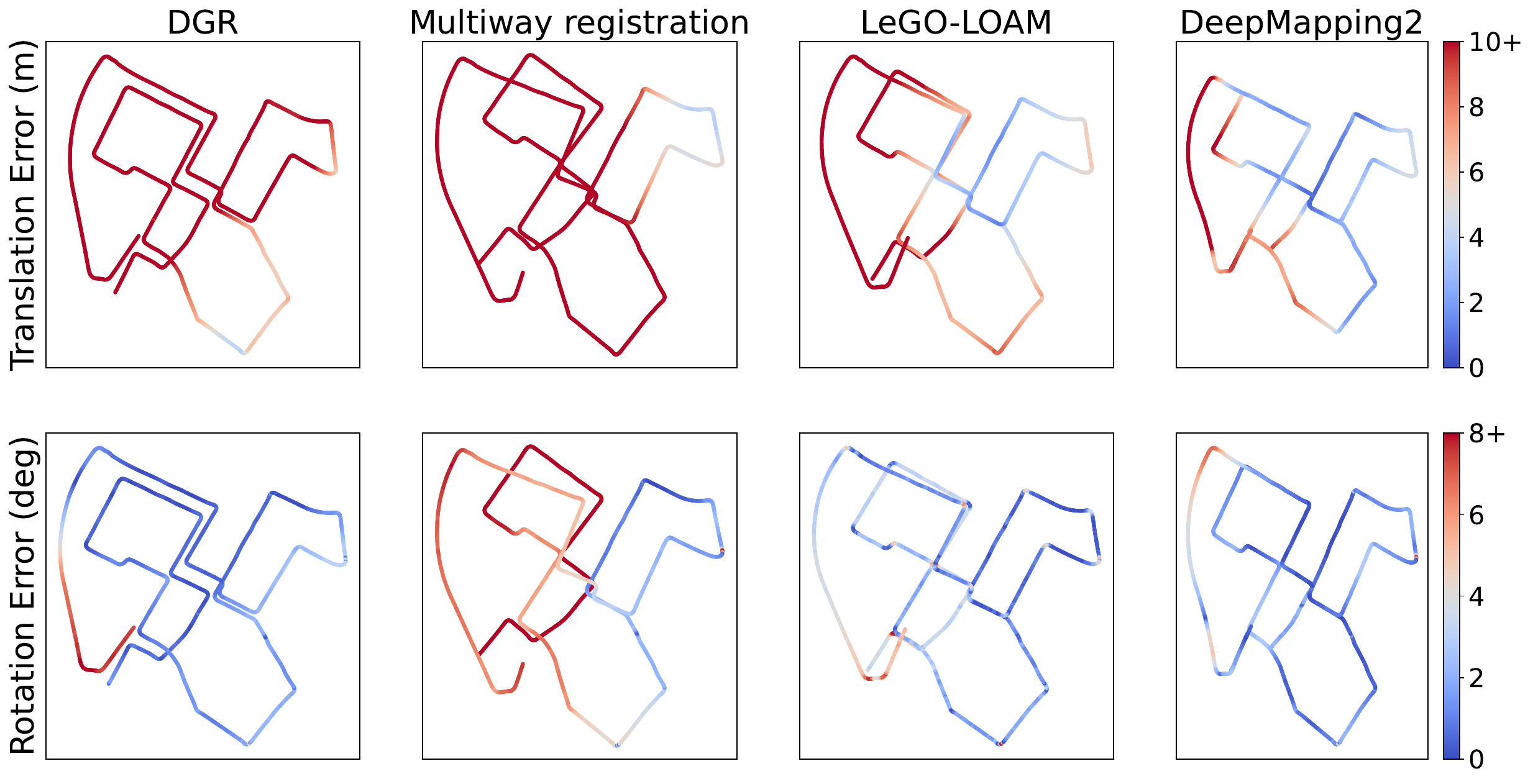}
    \caption{\textbf{KITTI drive\_0027.} Heat map visualization.}
    \label{fig:heat_kitti27}
\end{figure*}

\begin{figure*}[ht]
    \centering
    \includegraphics[width=1\textwidth]{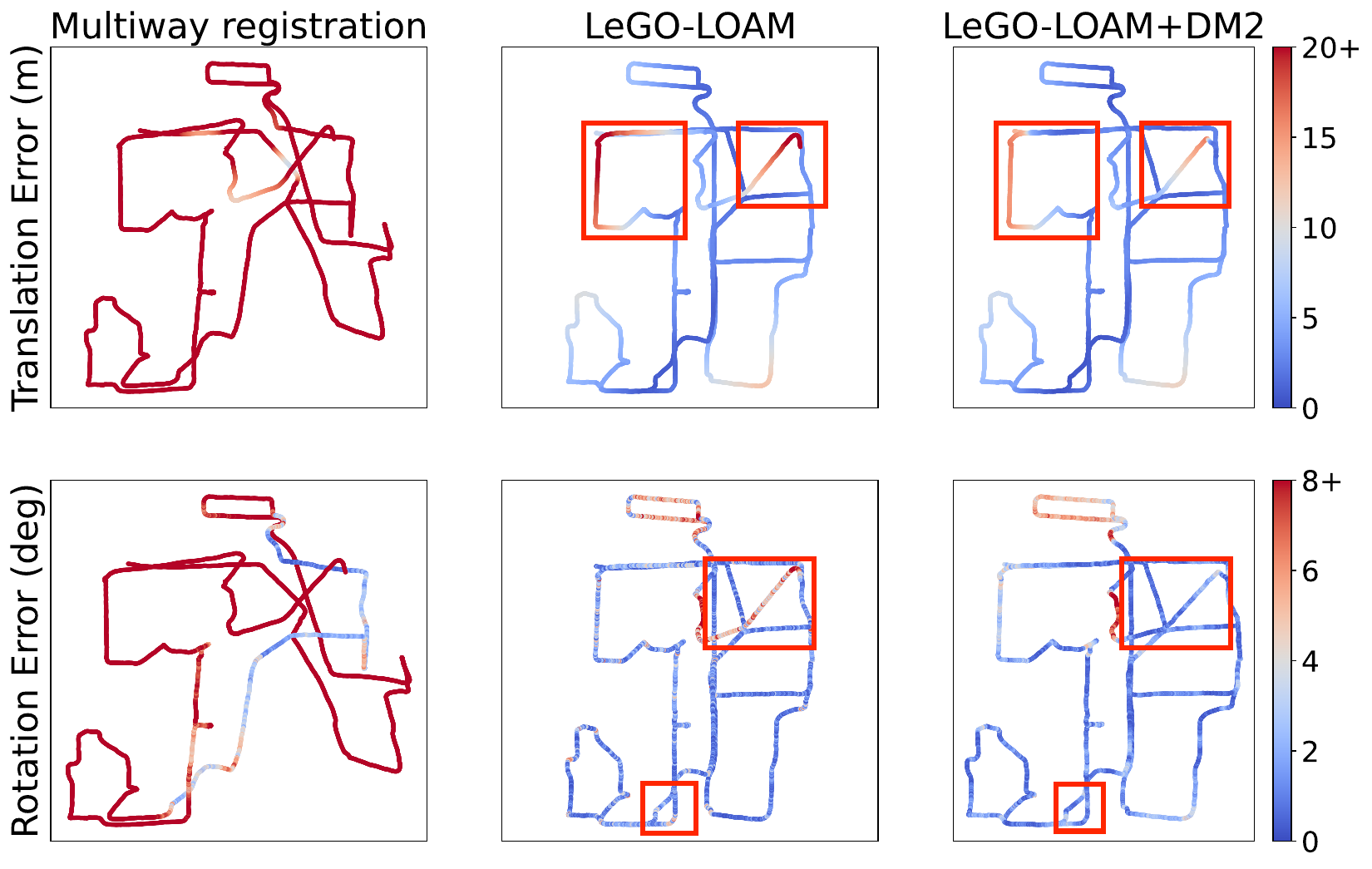}
    \caption{\textbf{NCLT.} Heat map visualization. The red boxes highlights the regions where DM2 improves over LeGO-LOAM.}
    \label{fig:heat_nclt}
\end{figure*}

In order to clearly demonstrate the optimization capability of DeepMapping2, we also offer heat map visualization. Results from \texttt{drive\_0018}, \texttt{drive\_0027}, and NCLT are included. It can be shown from \cref{fig:heat_kitti18,fig:heat_kitti27} that DeepMapping2 generally has smaller errors compared to other methods. Also, \cref{fig:heat_nclt} demonstrates how DeepMapping2 improves map from LeGO-LOAM, particularly for the areas indicated by the red box.

\begin{figure*}[ht]
    \centering
    \begin{subfigure}{0.32\textwidth}
        \includegraphics[width=1\textwidth]{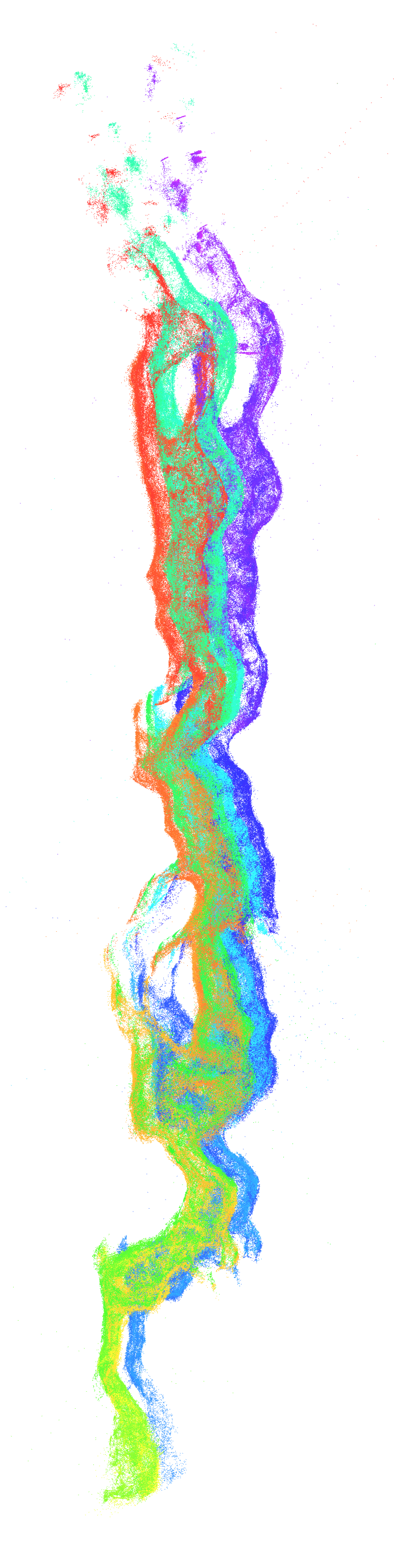}
        \caption{Kinematic odometry (KO)}
    \end{subfigure}
    \begin{subfigure}{0.32\textwidth}
        \includegraphics[width=1\textwidth]{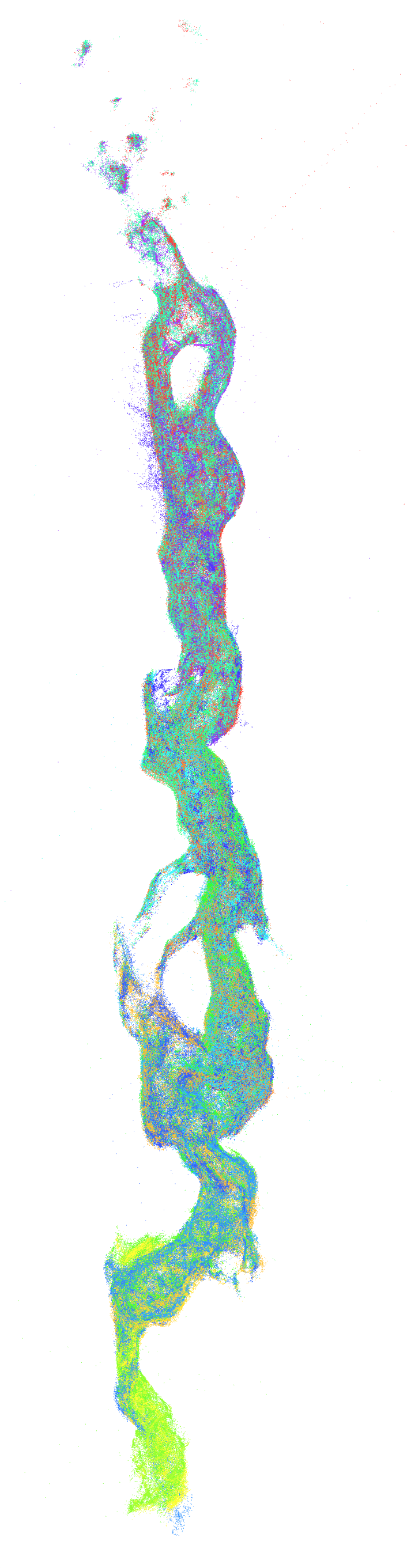}
        \caption{KO+DeepMapping2}
    \end{subfigure}
    \caption{\textbf{Mapping result on NeBula.} The color of point indicates the frame index in the trajectory.}\label{fig:supp_nebula}
\end{figure*}

We also include a larger and clearer visualization in \cref{fig:supp_nebula} for the mapping result on the NeBula dataset. It is clear that kinematic odometry fails to align the same locations when they are visited at different times, particularly at two ends of the map. On the other hand, DeepMapping2 significantly improves the map's quality.

\end{document}